\pdfoutput=1
\documentclass[11pt]{article}

\usepackage[preprint]{acl}

\usepackage{times}
\usepackage{latexsym}

\usepackage[T1]{fontenc}

\usepackage[utf8]{inputenc}

\usepackage{microtype}

\usepackage{inconsolata}

\usepackage{graphicx}

\usepackage{tcolorbox}
\usepackage{microtype}
\usepackage{url}
\usepackage{booktabs}
\usepackage{subfigure}

\usepackage{enumitem}
\newenvironment{itemize*}%
 {\leftmargini=20pt\begin{itemize}%
  \setlength{\itemsep}{3pt}%
  \setlength{\parskip}{0pt}%
  }%
 {\end{itemize}} 
\newenvironment{enumerate*}%
 {\begin{enumerate}%
  \setlength{\itemsep}{0pt}%
  \setlength{\parskip}{0pt}}%
 {\end{enumerate}}

\usepackage{amsmath}
\usepackage{cleveref}
\usepackage{listings}
\usepackage{titlesec}
\usepackage{multirow}
\usepackage{makecell}
\usepackage{subcaption} 
\usepackage{xcolor}

\usepackage{cuted}

\usepackage{array}       
\usepackage{caption}     
\usepackage{amssymb}
\usepackage{dsfont}

\definecolor{midnightgreen}{rgb}{0.0, 0.29, 0.33}
\definecolor{deepgreen}{HTML}{0aa344}
\definecolor{deeppurple}{HTML}{7030a0}
\definecolor{deepblue}{HTML}{171d91}
\definecolor{brown}{HTML}{843c0c}
\definecolor{shadered}{HTML}{ffe5e5}
\definecolor{shadegreen}{HTML}{e5f7ed}
\definecolor{msftBlack}{RGB}{0,0,0}
\definecolor{lightred}{RGB}{255,163,163}
\definecolor{deepred}{RGB}{146,0,0}

\newcommand{\red}{\textcolor{red}}
\newcommand{\green}{\textcolor{deepgreen}}

\usepackage{pifont}
\newcommand{\cmark}{\textcolor[rgb]{0.0, 0.6, 0.0}{\ding{51}}} 
\newcommand{\xmark}{\textcolor[rgb]{0.7, 0.0, 0.0}{\ding{55}}} 
\newcommand{\gmark}{\textcolor[rgb]{1,0.647,0}{\ding{51}}}

\NewDocumentCommand{\heng}
{ mO{} }{\textcolor{red}{\textsuperscript{\textit{Heng}}\textsf{\textbf{\small[#1]}}}}
\NewDocumentCommand{\cheng}
{ mO{} }{\textcolor{orange}{\textsuperscript{\textit{Cheng}}\textsf{\textbf{\small[#1]}}}}

\newcommand{\gymname}{UserBench}

\title{{\gymname}: An Interactive Gym Environment for User-Centric Agents}

\author{
Cheng Qian$^{1,2}$, Zuxin Liu$^{1}$, Akshara Prabhakar$^{1}$, Zhiwei Liu$^{1}$, Jianguo Zhang$^{1}$,\\
\textbf{Haolin Chen$^{1}$, Heng Ji$^{2}$, Weiran Yao$^{1}$, Shelby Heinecke$^{1}$, Silvio Savarese$^{1}$,}\\
\textbf{Caiming Xiong$^{1}$, Huan Wang$^{1}$}\vspace{5pt}\\
$^{1}$Salesforce AI Research \hspace{2pt} $^{2}$University of Illinois Urbana-Champaign \hspace{2pt} \\
}

\begin{document}
\maketitle
\begin{abstract}
Large Language Models (LLMs)-based agents have made impressive progress in reasoning and tool use, enabling them to solve complex tasks. However, their ability to proactively collaborate with users, especially when goals are vague, evolving, or indirectly expressed, remains underexplored. 
To address this gap, we introduce \textbf{{\gymname}}, a user-centric benchmark designed to evaluate agents in multi-turn, preference-driven interactions.
{\gymname} features simulated users who start with underspecified goals and reveal preferences incrementally, requiring agents to proactively clarify intent and make grounded decisions with tools. 
Our evaluation of leading open- and closed-source LLMs reveals a significant disconnect between task completion and user alignment. For instance, models provide answers that fully align with all user intents only 20\% of the time on average, and even the most advanced models uncover fewer than 30\% of all user preferences through active interaction. These results highlight the challenges of building agents that are not just capable task executors, but true collaborative partners. {\gymname} offers an interactive environment to measure and advance this critical capability.
All code and data are publicly available to support future research.\footnote{\ {\gymname} released at \url{https://github.com/SalesforceAIResearch/UserBench}}
\end{abstract}

\section{Introduction}

Large Language Models (LLMs) have demonstrated remarkable capabilities in complex reasoning~\citep{jaech2024openai, guo2025deepseek}, code generation~\citep{hui2024qwen2, guo2024deepseek,zhang2024diversity}, and solving advanced mathematical problems~\citep{yang2024qwen2, shao2024deepseekmath, qian2025modelingagent}. Their limitations in updated knowledge and precise computation are increasingly mitigated through tool use~\citep{liu2024apigen, qian2024investigate, zhang2024xlam, prabhakar2025apigen} and tool creation~\citep{qian2023creator, cai2023large, yuan2024craft, qian2024toolink}, enabling them to retrieve information and interact with external environments including database, web and games~\citep{deng2024mind2web, qian2024escapebench, zhu2025multiagentbench}. This tool-augmented reasoning grants LLMs agentic capabilities: they can autonomously execute tasks via these tools.

However, existing agentic environments often overlook a critical dimension: the role of the user. Despite achieving strong task performance, agents frequently fail to satisfy real user needs due to their inability to understand, adapt to, and collaborate with the task initiator~\citep{qian2024tell, lu2025proactive}. Current evaluations primarily assess tool use and task execution, rarely considering whether the agent effectively interprets and aligns with the user’s underlying and evolving intent.

This motivates our central research question: \textbf{How can we evaluate agents from a user-centric perspective?} To answer this, we first examine how users typically communicate goals. Human communication is inherently a joint activity, where meaning is co-constructed through interaction \citep{clark1996using}. Moreover, language is inherently ambiguous, making it difficult for users to fully and clearly convey their intent in a single interaction.~\citep{liu2023we}. As such, user instructions tend to share three core traits:
(i) \textbf{Underspecification}: users often initiate requests before fully formulating their goals;
(ii) \textbf{Incrementality}: intent emerges and evolves across interaction; and
(iii) \textbf{Indirectness}: users may obscure or soften their true intent due to social or strategic reasons.

Revolving around these traits, we introduce \textbf{{\gymname}}, a user-centric environment designed to facilitate an agent’s ability to engage in meaningful, multi-turn interactions with users who exhibit these traits. In {\gymname}, simulated users provide initial vague task instruction (underspecification), gradually reveal preferences over time (incrementality), and often do so implicitly (indirectness). Agents must proactively clarify goals, interpret subtle cues, and adaptively reason through tool use to succeed.

Built on the standard Gymnasium framework, {\gymname} offers a modular, extensible setup with a standardized interaction interface and a stable tool-use backend, enabling rigorous and reproducible evaluation. We benchmark several leading open- and closed-source models and find that current LLMs still struggle to interactively uncover and act on user preferences.
For instance, scores drop by over 40\% on average when models are restricted to selecting only one option per traveling aspect in {\gymname}, revealing their difficulty in making optimal decisions. Moreover, models provide answers that fully align with all user intents only 20\% of the time on average, and even the best-performing models elicit less than 30\% of all user preferences through active querying, suggesting limited ability to engage in purposeful, user-driven dialogue. While strong models handle tool use reliably, they remain brittle in understanding implicit and nuanced human needs, highlighting the core challenge that {\gymname} aims to measure. We summarize our contributions as follows:
\begin{itemize}[topsep=2pt, partopsep=-3pt, leftmargin=8pt, itemsep=-3pt]
  \item \textbf{Data}: We propose 4K+ scenarios that capture grounded communication challenges (underspecification, incrementality, indirectness) through a carefully curated pipeline.
  \item \textbf{Environment}: We introduce {\gymname}, a scalable and modular gym environment designed both as a benchmark and a training ground for LLM agents engaged in multi-turn, preference-driven user interactions.
  \item \textbf{Analysis}: We show that despite strong tool use capability, current models still struggle to comprehensively uncover and align with user intent, revealing limits in user-centric reasoning.
\end{itemize}

We view this work as a foundational step toward truly user-centric agents: not just efficient executors, but collaborative teammates capable of aligning with nuanced human intent.

\section{Related Work}

\paragraph{User-centric environments for LLM evaluation.}  
Recent work has increasingly emphasized the importance of evaluating LLMs under realistic, user-centric conditions. Traditional benchmarks assess task success assuming fully specified prompts, overlooking how users often communicate underspecified, incremental, or indirect goals. To address this, benchmarks like User Reported Scenarios \citep{wang2024user}, Intention-in-Interaction \citep{qian2024tell}, and WildBench \citep{lin2024wildbench} compile real user queries to evaluate whether models align with nuanced human intent. These datasets emphasize preference satisfaction, multi-intent understanding, and performance in ``in-the-wild'' conditions. Complementing this, other benchmarks such as MINT \citep{wang2024mint}, PrefEval \citep{zhao2025prefeval}, $\tau$-Bench \citep{yao2024tau}, and $\tau^{2}$-Bench~\citep{barres2025tau} focus on dynamic, multi-turn interactions, testing whether agents can incorporate feedback, handle evolving preferences, and maintain user alignment over time. However, several limitations remain. For example, a large portion of interactions are automatically synthesized in these benches, often resulting in overly lengthy and unnatural goal formulations that diverge from how users typically express themselves. In contrast, {\gymname} explicitly models the three core interactive traits of user communication that are critical for real-world alignment. It also spans hundreds of diverse goal configurations, enabling fine-grained evaluation on a wide range of user intents, thus providing a more comprehensive and behaviorally grounded setting for assessing user-centric agents. We provide a comprehensive comparison of our gym environment with recent environments and benchmarks in \Cref{tab:comparison}.

\begin{table*}[!t]
\centering
\resizebox{\linewidth}{!}{
\begin{tabular}{lccccccccc}
\toprule
\textbf{Benchmark} 
& \textbf{\makecell{Multi-turn\\Interaction}} 
& \textbf{\makecell{Goal\\Ambiguity}} 
& \textbf{\makecell{Tool\\Use}} 
& \textbf{\makecell{Dynamic\\State}} 
& \textbf{\makecell{Multi-Aspect\\Reasoning}} 
& \textbf{\makecell{User\\Simulation}} 
& \textbf{\makecell{Domain\\Diversity}} 
& \textbf{\makecell{Custom-\\izable}} 
& \textbf{\makecell{Scal-\\able}} \\
\midrule
\textit{$\tau^2$-Bench\citep{barres2025tau}}          
& \cmark  
& \xmark  
& \cmark  
& \cmark  
& \xmark  
& \cmark  
& \xmark  
& \gmark  
& \gmark  
\\
\textit{$\tau$-Bench\citep{yao2024tau}}                
& \cmark  
& \xmark  
& \cmark  
& \cmark  
& \gmark  
& \cmark  
& \gmark  
& \gmark  
& \gmark  
\\
\textit{ToolSandbox\citep{lu2024toolsandbox}}              
& \cmark  
& \gmark  
& \cmark  
& \cmark  
& \gmark  
& \cmark  
& \gmark  
& \gmark  
& \gmark  
\\
\textit{MINT\citep{wang2024mint}}                           
& \cmark  
& \gmark  
& \cmark  
& \cmark  
& \gmark  
& \cmark  
& \gmark  
& \gmark  
& \gmark  
\\
\textit{IN3\citep{qian2024tell}}  
& \cmark  
& \cmark  
& \xmark  
& \xmark  
& \xmark  
& \cmark  
& \gmark  
& \xmark  
& \xmark  
\\
\textit{ToolTalk\citep{farn2023tooltalk}}                   
& \gmark  
& \xmark  
& \cmark  
& \xmark  
& \xmark  
& \xmark  
& \gmark  
& \xmark  
& \xmark  
\\
\textit{API-Bank\citep{li2023apibank}}                      
& \xmark  
& \xmark  
& \cmark  
& \xmark  
& \xmark  
& \xmark  
& \gmark  
& \gmark  
& \gmark  
\\
\textit{MetaTool\citep{huang2023metatool}}                  
& \xmark  
& \xmark  
& \cmark  
& \xmark  
& \xmark  
& \xmark  
& \gmark  
& \gmark  
& \gmark  
\\
\midrule
\textbf{{\gymname} (Ours)}              
& \cmark  
& \cmark  
& \cmark  
& \cmark  
& \cmark  
& \cmark  
& \cmark  
& \cmark  
& \cmark  
\\
\bottomrule
\end{tabular}
}
\caption{For each existing benchmark, the table indicates whether the corresponding trait is fully addressed (\cmark), partially addressed (\gmark), or not addressed (\xmark). We provide detailed explanation for each trait being compared in \Cref{sec:apdx_comparison}.}
\label{tab:comparison}
\end{table*}

\paragraph{User-centric agent designs.}  
Designing agents that collaborate effectively with users requires modeling ambiguity, evolving intent, and user-specific preferences. Standard instruction-tuned models often hallucinate intent or avoid asking clarifying questions. Recent work aims to teach agents to proactively clarify underspecified instructions. For instance, models trained with simulated ambiguity resolution \citep{zhang2024clarify, chen2025act} better recognize when to ask versus answer. Other work tackles user modeling directly: TravelPlanner+ \citep{singh2024personal} and PRELUDE \citep{gao2024prelude} build agents that personalize responses using either explicit profiles or latent preferences inferred from user edits. While these designs make progress toward adaptive, user-aware agents, they often target narrow personalization tasks or rely on static user models; in contrast, our work provides a general-purpose evaluation environment that systematically tests an agent’s ability to dynamically uncover and respond to emergent user intent across varied and customizable interaction scenarios.

\begin{figure*}
    \centering
    \includegraphics[width=\linewidth]{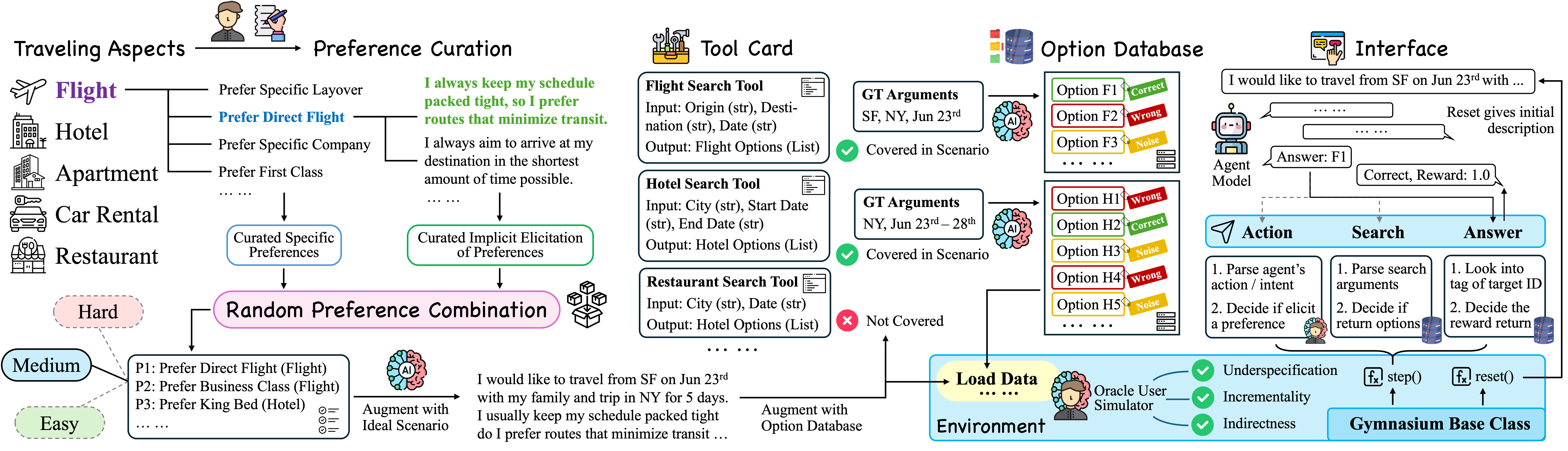}
    \caption{The pipeline of {\gymname} Construction, including data gathering, preference curation, tool augmentation, environment setup and interface design. \textit{Random Preference Combinations} makes our pipeline highly scalable and enables the efficient generation of even 10K+ data points.} 
    \label{fig:pipeline}
\end{figure*}

\section{{\gymname}}

We build {\gymname} upon the standardized gymnasium environment, focusing on travel planning tasks where users gradually reveal their preferences in implicit ways. Compared to prior work, {\gymname} features a significantly more diverse and grounded collection of user preferences, each paired with carefully curated implicit expressions, and supports tool-augmented search with injected noise and cost-awareness. This setting challenges agents to perform search and reasoning in a user-centric manner, under constraints of partial, evolving, and indirect preference information. We describe the gym design details in the following.

\subsection{Data Gathering}

As shown in \Cref{fig:pipeline}, we begin by identifying five core aspects of travel planning: flight, hotel, apartment, car rental, and restaurant. Our data gathering is guided by three key principles: \textit{realism}, \textit{diversity}, and \textit{implicitness}. First, we ensure realism by curating preferences that reflect plausible, real-world user needs—such as preferring direct flights (flight) or king-sized beds (hotel). Second, we aim for diversity by collecting around a hundred distinct preferences that span travel aspects. Third, we emphasize implicitness by pairing each preference with multiple naturally phrased statements that express the intent indirectly. For instance, the preference for direct flights might be expressed as: ``I always keep my schedule packed tight, so I prefer travel routes that minimize transit time.'' These implicit expressions are carefully curated and validated to ensure that (i) they do not reveal the preference explicitly, and (ii) they still clearly imply the intended preference without ambiguity.

These curated preferences and their associated implicit expressions serve as the foundation for generating travel scenarios. We randomly sample and combine preferences across the five aspects, categorizing the resulting data into three difficulty tiers based on the number of preferences involved. For each combination, we generate an ideal travel arrangement statement that incorporates all relevant preferences using GPT-4o. This finally forms a data point in our environment.

\subsection{Tool Augmentation}
For each travel planning aspect, we develop a corresponding tool that simulates a database search. For example, the flight search tool accepts origin, destination, and date, returning a list of mock options. Instead of querying real-time data, we rely on pre-generated options to ensure stable and controlled tool outputs. This design choice is justified because (i) it guarantees consistency and quality of returned results, and (ii) our focus is user-centric reasoning rather than challenges of online search.

Each data point is associated with a tailored database of options for all relevant travel planning aspects. Specifically, for each aspect, we generate three types of options: correct (satisfying all preferences), wrong (violating at least one preference), and noise (incomplete or irrelevant to the query, such as a flight option with a destination different from the one the user searched for). These options are mixed to form the complete search space, with each data point containing up to over 100 options. Option generation is handled by GPT-4o under human supervision to ensure quality and diversity.

\subsection{Environment Building}
Each data point in the environment contains a travel scenario, a set of implicit user preferences, and a database of relevant options. The environment simulates an oracle user with access to these internal states, while engaging in multi-turn interactions with a agent model under test. Upon reset, the environment provides only basic trip information and high-level requirements (e.g., hotel or flight), without disclosing any of its specific preferences. Preferences are then elicited over time, either (i) when the tested model explicitly asks about a relevant aspect, or (ii) after a fixed number of turns without progress (see \Cref{sec:apdx_construction} for details), at which point a preference is randomly and proactively revealed. All elicited preferences are presented implicitly, in line with the associated data.

This design captures the three key traits and challenges we previously identify: \textit{underspecification}, where only coarse needs are initially provided; \textit{incrementality}, where user preferences emerge progressively as the conversation evolves; and \textit{indirectness}, where preferences are stated implicitly through natural, carefully-curated, heuristic-based language. These properties reflect real-world user behaviors and make {\gymname} a robust testbed for user-centric reasoning.

\subsection{Agent Interaction}
We provide a standardized interaction interface for agents, supporting three types of actions: \texttt{action}, \texttt{search}, and \texttt{answer}. Through \texttt{search}, agents issue queries specifying the travel planning aspect and arguments (e.g., hotel with specific dates), which are matched against ground truth. If the arguments match, the environment returns the mixed set of relevant options; otherwise, it returns nothing. Through \texttt{action}, the agent communicates with the user, typically by asking clarifying questions. The environment interprets the intent and responds naturally, potentially revealing a preference in an implicit form. Finally, through \texttt{answer}, the agent selects its recommended option ID(s), which are evaluated against the ground-truth correct options. Since ideal itineraries often span multiple aspects, correct answers may include several options.

\begin{table*}[!t]
\begin{center}
\small
\tabcolsep=0.015\linewidth
\resizebox{\linewidth}{!}{
\begin{tabular}{lccll|lll|lcc}
\toprule
\textbf{Difficulty Tier} & \textbf{\# Test} & \textbf{\# Train} & \textbf{Composition} & & \multicolumn{2}{c}{\textbf{Global Dataset Metadata}} & & \textbf{Travel Aspect} & \textbf{\# Scenarios} & \textbf{Preferences} \\
\midrule
Easy   & 118 & 666 & Travel-22, 2222   & & Travel Aspects       & 5   &                         & Hotel       & 187 (39.7\%) & 15 \\
Medium & 201 & 1130 & Travel-33, 233, 333   & & Pref. Categories     & 24  &                         & Restaurant  & 298 (63.3\%) & 19 \\
Hard   & 152 & 855 & Travel-44, 334, 444 & & Total Prefs.         & 82  &                         & Flight      & 275 (58.4\%) & 16 \\
\multicolumn{4}{l}{-----------------------------------------------------------------------------} & & Elicitation Ways     & 902 &                         & Rental Car  & 243 (51.6\%) & 14 \\
\textbf{Total} & \textbf{471} & 2651 & \textbf{{\gymname} Test} / Train &  & Options (B/C/W/N)     & 1/3/10/5 &                   & Apartment    & 172 (36.5\%) & 18 \\
\bottomrule
\end{tabular}
}
\end{center}
\vspace{-3mm}
\caption{\textbf{{\gymname} Statistics:} Difficulty-tiered statistics (left), global configuration metadata (middle), and travel aspect coverage (right). \textit{Pref.} denotes preference; \textit{B/C/W/N} stands for Best, Correct, Wrong, and Noise options respectively; \textit{\# Scenarios} indicates the number of data scenarios in which each travel aspect appears.}
\label{tab:statistics}
\end{table*}

\subsection{Variants, Scalability and Statistics}
\label{sec:variants}
{\gymname} supports several extensions. First, it can simulate a noisy search environment where query results are corrupted at a configurable frequency, mimicking real-world web or tool errors. Second, it allows for budget-aware decision-making by rewarding models that not only find correct options but also select the most cost-effective ones (best option). A hyperparameter can control the strength of this reward, encouraging agents to optimize toward practical utility.

As illustrated in \Cref{fig:pipeline}, the Random Preference Combination strategy enhances both the diversity and scalability of our data construction, enabling over 10K distinct scenarios. To manage computational costs during benchmarking, we use 417 data points for testing and reserve 2651 for training, which are not used in benchmarking but may support future model training (see Discussion section). Detailed statistics of {\gymname} are presented in \Cref{tab:statistics}. For instance, ``Travel-223'' denotes a scenario with three travel aspects: two containing two implicit preferences each, and one with three. Additional details on data and environment construction are provided in the \Cref{sec:apdx_construction}.

\section{Experiments}

\begin{table*}[!t]
\begin{center}
\small
\tabcolsep=0.015\linewidth
\resizebox{\linewidth}{!}{
\begin{tabular}{lcccccc}
\toprule
\textbf{Model Name} & \textbf{Best Exist Rate} & \textbf{Correct Exist Rate} & \textbf{Score} & \textbf{\makecell{Search Attempt\\Valid (\%)}} & \textbf{\makecell{Action Attempt\\Valid (\%)}} & \textbf{\makecell{Preference Elicited (\%)\\(Active / Passive)}} \\
\midrule
GPT-4o & 0.204 & \textbf{0.361} & \textbf{0.329} & 82.48 & 27.82 & 27.32 (24.06 / 3.26) \\
Gemini-2.5-Pro & 0.245 & 0.328 & 0.317 & 77.40 & 29.26 & 29.71 (23.85 / 5.85) \\
Claude-4-Sonnet & \textbf{0.260} & 0.318 & 0.307 & 74.28 & 24.26 & \textbf{34.25} (\textbf{26.31} / 7.94) \\
Deepseek-V3 & 0.148 & 0.218 & 0.210 & 61.22 & 23.62 & 31.82 (18.90 / 12.92) \\
Qwen3-14B & 0.157 & 0.197 & 0.209 & 70.56 & \textbf{33.71} & 30.07 (15.95 / 14.12) \\
Qwen3-32B & 0.154 & 0.211 & 0.206 & 79.42 & 26.89 & 22.01 (13.04 / 8.96) \\
Llama-3.3-70B & 0.120 & 0.209 & 0.198 & 60.51 & 11.29 & 23.25 (13.05 / 10.20) \\
Qwen3-8B & 0.140 & 0.186 & 0.180 & 68.94 & 23.20 & 28.86 (11.15 / 17.70) \\
GPT-4o-mini & 0.102 & 0.173 & 0.166 & 76.10 & 13.55 & 25.43 (12.91 / 12.52) \\
Llama-3.1-8B & 0.073 & 0.171 & 0.159 & 57.99 & 12.13 & 19.87 (8.09 / 11.78) \\
Gemini-2.5-Flash & 0.047 & 0.080 & 0.125 & \textbf{83.62} & 28.14 & 14.00 (11.15 / 2.86) \\
\bottomrule
\end{tabular}
}
\end{center}
\vspace{-3mm}
\caption{{\gymname} main evaluation results across different models (single-choice setting).}
\label{tab:travelgym_results_onechoice}
\end{table*}

\begin{table*}[!t]
\begin{center}
\small
\tabcolsep=0.015\linewidth
\resizebox{\linewidth}{!}{
\begin{tabular}{lcccccc}
\toprule
\textbf{Model Name} & \textbf{Best Exist Rate} & \textbf{Correct Exist Rate} & \textbf{Score} & \textbf{\makecell{Search Attempt\\Valid (\%)}} & \textbf{\makecell{Action Attempt\\Valid (\%)}} & \textbf{\makecell{Preference Elicited (\%)\\(Active / Passive)}} \\
\midrule
GPT-4o & \textbf{0.652} & \textbf{0.725} & \textbf{0.710} & 85.96 & 37.89 & 15.10 (13.11 / 1.99) \\
Gemini-2.5-Pro & 0.604 & 0.685 & 0.673 & 80.43 & 38.01 & 37.25 (28.80 / 8.46) \\
Claude-4-Sonnet & 0.586 & 0.619 & 0.612 & \textbf{87.94} & \textbf{41.17} & \textbf{42.21} (\textbf{30.67} / 11.54) \\
Qwen3-32B & 0.343 & 0.416 & 0.411 & 82.76 & 34.24 & 17.72 (10.81 / 6.90) \\
Deepseek-V3 & 0.351 & 0.398 & 0.391 & 61.91 & 21.06 & 16.96 (10.18 / 6.78) \\
Llama-3.3-70B & 0.294 & 0.366 & 0.372 & 63.46 & 21.72 & 31.72 (16.68 / 15.04) \\
Gemini-2.5-Flash & 0.279 & 0.316 & 0.311 & 84.27 & 33.58 & 15.32 (11.67 / 3.64) \\
GPT-4o-mini & 0.227 & 0.294 & 0.283 & 81.54 & 24.90 & 20.18 (10.46 / 9.72) \\
Qwen3-14B & 0.204 & 0.249 & 0.258 & 71.48 & 35.06 & 26.21 (13.07 / 13.14) \\
Llama-3.1-8B & 0.117 & 0.252 & 0.234 & 61.38 & 12.20 & 26.01 (7.80 / 18.20) \\
Qwen3-8B & 0.147 & 0.186 & 0.182 & 69.83 & 23.26 & 20.31 (7.95 / 12.37) \\
\bottomrule
\end{tabular}
}
\end{center}
\vspace{-3mm}
\caption{{\gymname} main evaluation results across different models (multi-choice setting).}
\label{tab:travelgym_results_multichoice}
\end{table*}

\subsection{Settings}
\paragraph{Gym Settings.}
We adopt both variants (corrupted search, budget constraint) to increase test difficulty. GPT-4o with a temperature of 0.0 serves as the user simulator for all conversations. In the standard setup, we limit the maximum number of conversation turns to 20. We evaluate models under two settings: (1) a single-choice setting (our main evaluation), where the model is allowed to output only one option for each travel aspect, and (2) a multi-choice setting, where the model may output multiple options, and we evaluate it based on the option that achieves the highest reward.

\paragraph{Models.}
Our evaluation includes both closed-source and open-source models. Closed-source models include the GPT, Claude, Deepseek, and Gemini families, while open-source models include the Qwen3 and Llama3 families, with model sizes ranging from 8B to 70B. All models generate responses using a temperature of 0.0 to ensure deterministic behavior.

\paragraph{Metrics.}
\label{sec:metrics}
The main evaluation metric is a normalized \textbf{score} based on the quality of selected options for each aspect of a travel scenario. For each aspect, if the model selects the best option (as judged by reward), it receives a score of 1.0. If the option is correct but not the best, it receives 0.8. All other options receive 0.0. We compute the highest reward score among the options selected for each aspect (in the multi-choice setting) and average this over all aspects in the scenario. For example, if a scenario has two aspects, and the model selects options with rewards 1.0 and 0.8 for the first aspect, and 0.0, 0.8, and 0.0 for the second, the score is:

\begin{small}
\[
\text{Final Score} = \frac{\max(1.0, 0.8) + \max(0.0, 0.8, 0.0)}{2} = 0.9
\]
\end{small}

In the single-choice setting, only the first option per aspect is considered, so the max function does not apply and the score is averaged only for the first choice of each aspect. In addition, we also report several auxiliary metrics to better understand model behavior (all micro-averaged):
\begin{itemize}[topsep=2pt, partopsep=-3pt, leftmargin=8pt, itemsep=0pt]
\item \textbf{Best Exist Rate}: The proportion of aspects where the model includes the best option among its selected choices.
\item \textbf{Correct Exist Rate}: The proportion of aspects where the model includes one correct (not necessarily best) option.
\item \textbf{Valid Search Attempt (\%)}: The rate at which the model's search queries are syntactically valid.
\item \textbf{Valid Action Attempt (\%)}: The rate at which the model's actions successfully probe the user’s real preferences.
\item \textbf{Preference Elicited (\%)}: The percentage of all ground-truth preferences revealed during the conversation. It includes \textit{Active} elicitation, where preferences are disclosed in response to valid action attempts by the tested model, and \textit{Passive} elicitation, where preferences are released by the {\gymname} to guide the conversation when the model becomes too off-topic.
\end{itemize}
Please refer to \Cref{sec:apdx_exp_setting} for a more detailed explanation of each metric.

\subsection{Results}

We present the results for the two settings in~\Cref{tab:travelgym_results_onechoice} and \Cref{tab:travelgym_results_multichoice}, and summarize our key findings below.

\paragraph{Single-choice setting is significantly more challenging.}
On average, scores drop by approximately 40\% when switching from the multi-choice to the single-choice setting. This highlights the difficulty models face in selecting the best, or even one of the three correct, options with only one answer attempt. When allowed to propose multiple answers, we observe that the performance generally improves, indicating that having more chances increases the likelihood of hitting a correct answer.

\paragraph{Preference elicitation remains low across models.}  
One might expect that performance gains in the multi-choice setting stem from better user understanding. However, we find that preference elicitation rates do not significantly improve, and in some cases, such as GPT-4o and Deepseek-V3, even decline. This suggests that higher scores more often result from random guessing or brute-force coverage rather than active reasoning. Moreover, the overall user preference elicitation rate remains low across models, especially for preferences revealed through active queries. This indicates that current models still struggle to proactively and effectively uncover user needs in interactive settings.

\paragraph{Understanding users is harder than executing tool use.}  
Top-performing models maintain over 80\% success in valid search attempts, but the rate of valid action attempts is much lower. This aligns with {\gymname}’s goal of shifting difficulty from tool use to user understanding. Note that valid action attempt rate is computed only over action steps (excluding search and answer), so a higher rate does not necessarily reflect a higher absolute number of valid queries overall. Still, it indicates that the model’s questions are more precise and preference-relevant, rather than vague or off-topic, which we classify as invalid attempts.

\paragraph{Other Interesting Findings.} Some models, such as Gemini-2.5-Flash, achieve high Action Attempt Valid rates but relatively low Active Preference Elicited. This suggests they can ask effective clarifying questions but often fail to do so repeatedly or comprehensively, limiting their ability to capture the full scope of user preferences. Please see the \Cref{sec:apdx_exp_setting} for detailed metric explanations to help understanding this result. In contrast, Claude-4-Sonnet performs well on both metrics, indicating strong capability in intent clarification. However, its overall score is not the highest. This highlights a gap between understanding user preferences and effectively integrating them into decision-making. In other words, even when models excel at eliciting preferences, they may still struggle to leverage that information in their reasoning to deliver optimal recommendations.

\section{Analysis}
All analyses are conducted under the single-choice setting, which serves as our main evaluation setup, and focus on the \textit{score} as the primary metric. The turn-based, pass-$k$ sampling, and choice ablation analyses are performed specifically on data points from the Travel-22, 33, and 44 scenarios.

\begin{figure}[!t]
    \centering
    \begin{minipage}{1.0\linewidth}
        \centering
        \vspace{0pt}
        \includegraphics[width=\linewidth]{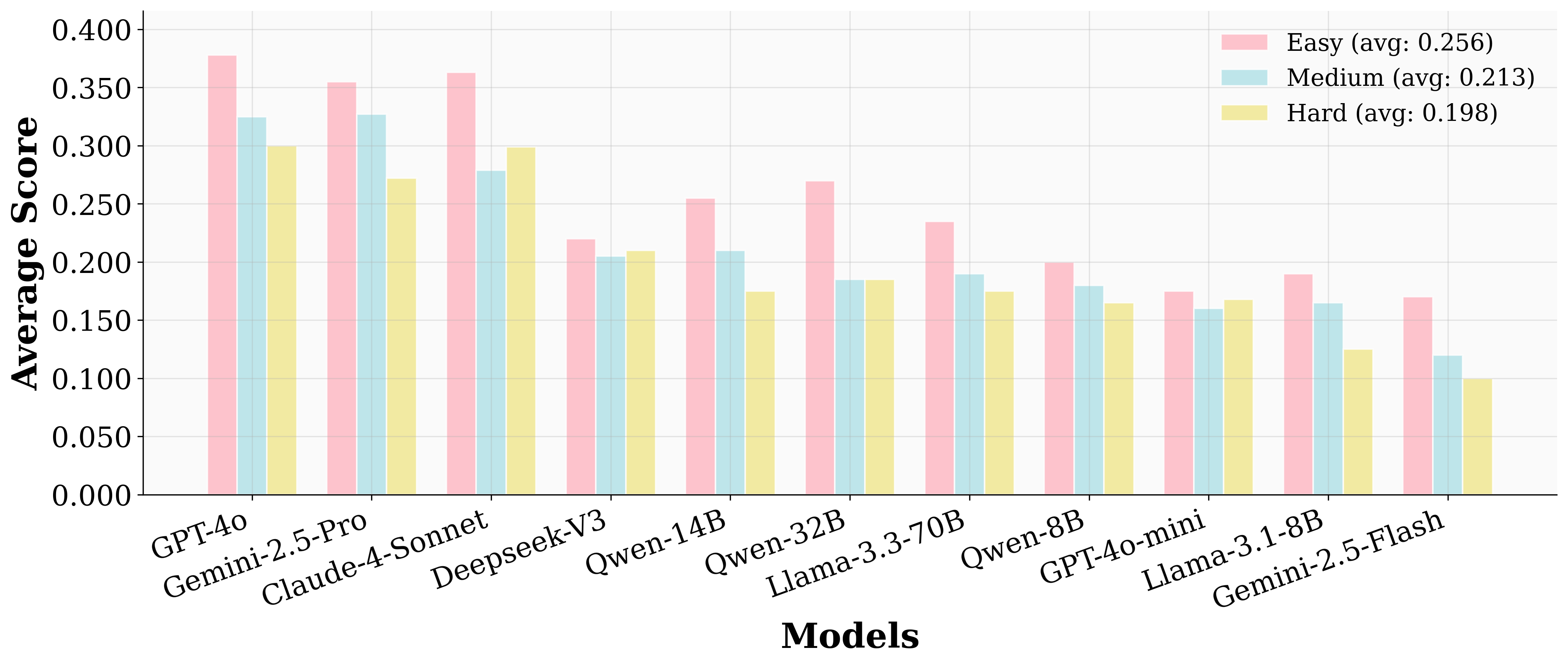}
    \end{minipage}
 
    \caption{The score distribution of tested models across three difficulty tiers.}
    \label{fig:difficulty}
    
    \vspace{0pt}
\end{figure}

\paragraph{Difficulty tier division effectively reflects {\gymname}'s challenges.}  
We begin by analyzing how scenario difficulty impacts model performance, as shown in~\Cref{fig:difficulty}. Following the tiering in~\Cref{tab:statistics}, we divide test scenarios into Easy, Medium, and Hard based on preference complexity. Results show a general downward trend in scores as difficulty increases. This confirms that our difficulty stratification captures real reasoning challenges for models. Furthermore, the performance drop within each model across tiers reveals a lack of robustness in managing complex user interactions, which is something humans typically excel at.

\begin{figure}[!t]
    \centering
    \begin{minipage}{1.0\linewidth}
        \centering
        \vspace{0pt}
        \includegraphics[width=\linewidth]{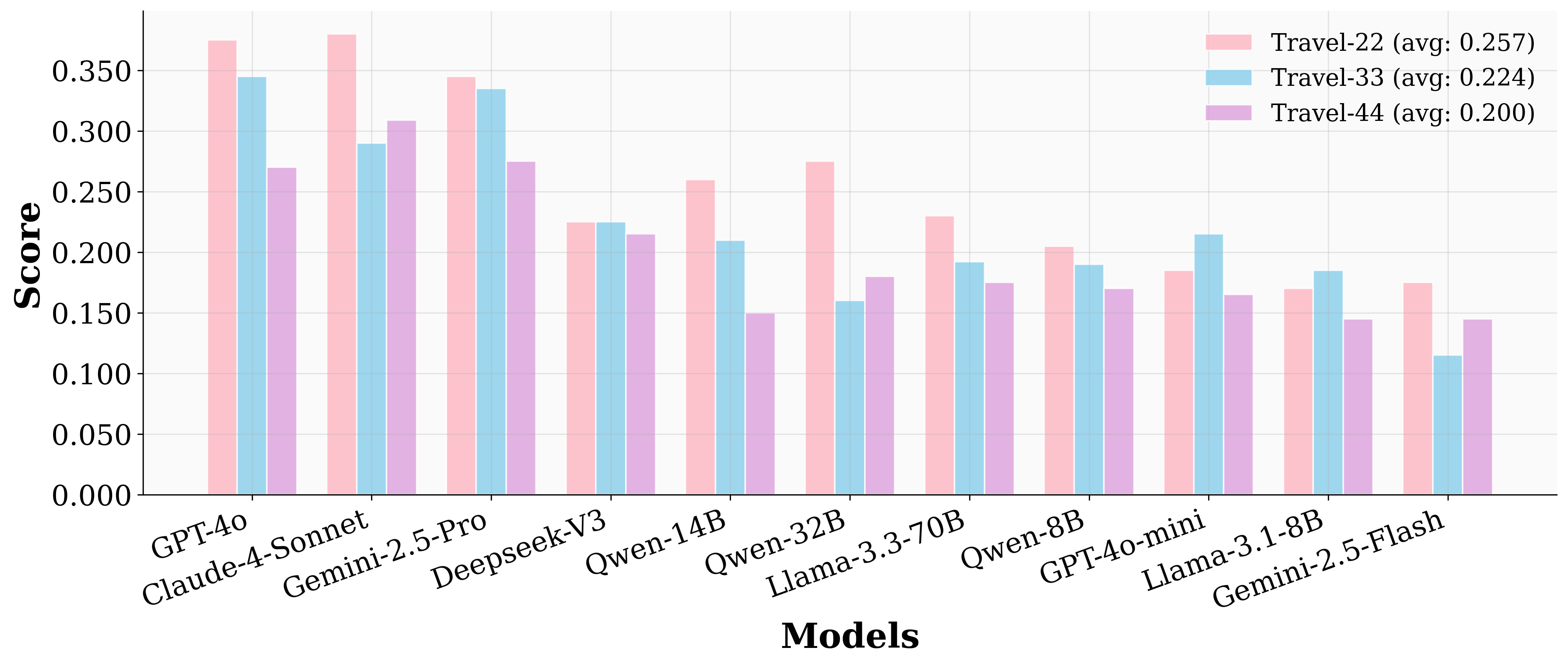}
    \end{minipage}

    \vspace{2pt}
    
    \begin{minipage}{1.0\linewidth}
        \centering
        \vspace{0pt}
        \includegraphics[width=\linewidth]{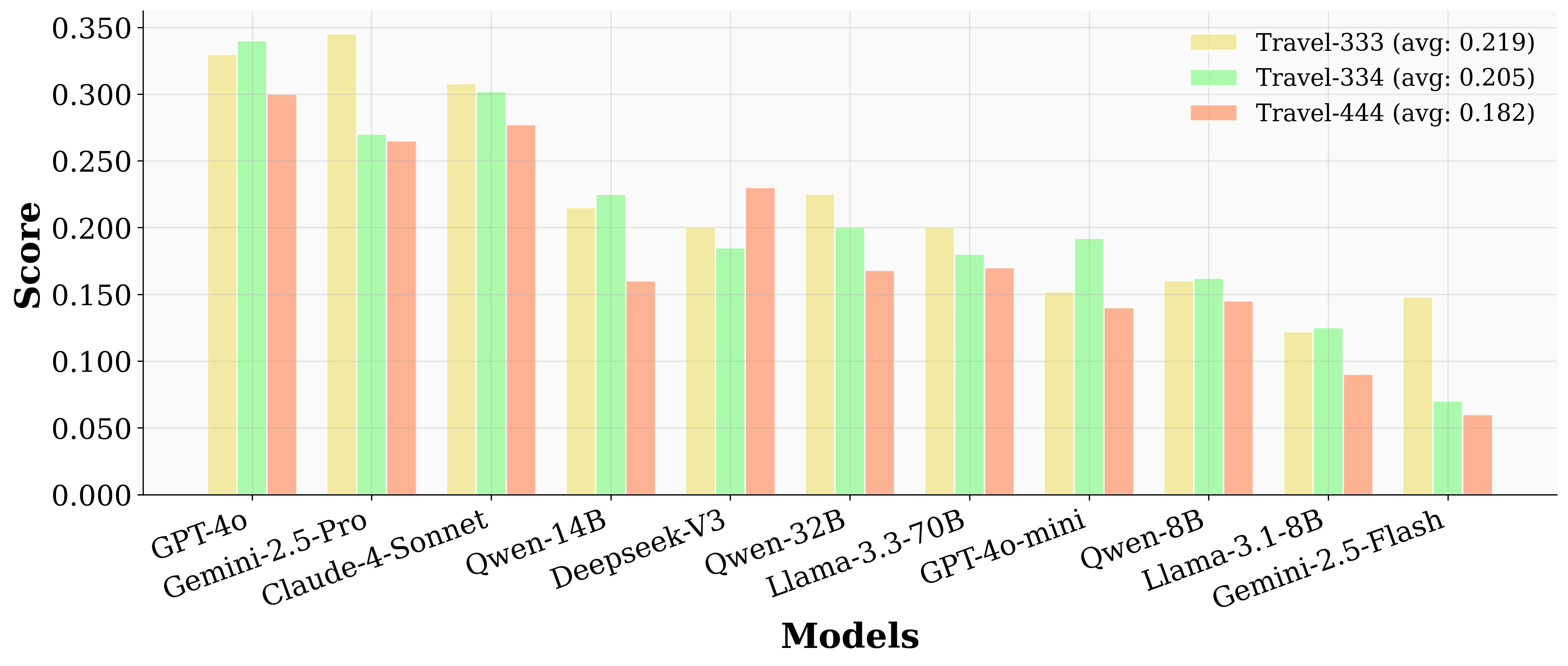}
    \end{minipage}
    
    \caption{When aspect number is settled, more user preferences generally lead to less scores.}
    \label{fig:preference}
    
    \vspace{0pt}
\end{figure}

\begin{figure}[!t]
    \centering
    \begin{minipage}{1.0\linewidth}
        \centering
        \vspace{0pt}
        \includegraphics[width=\linewidth]{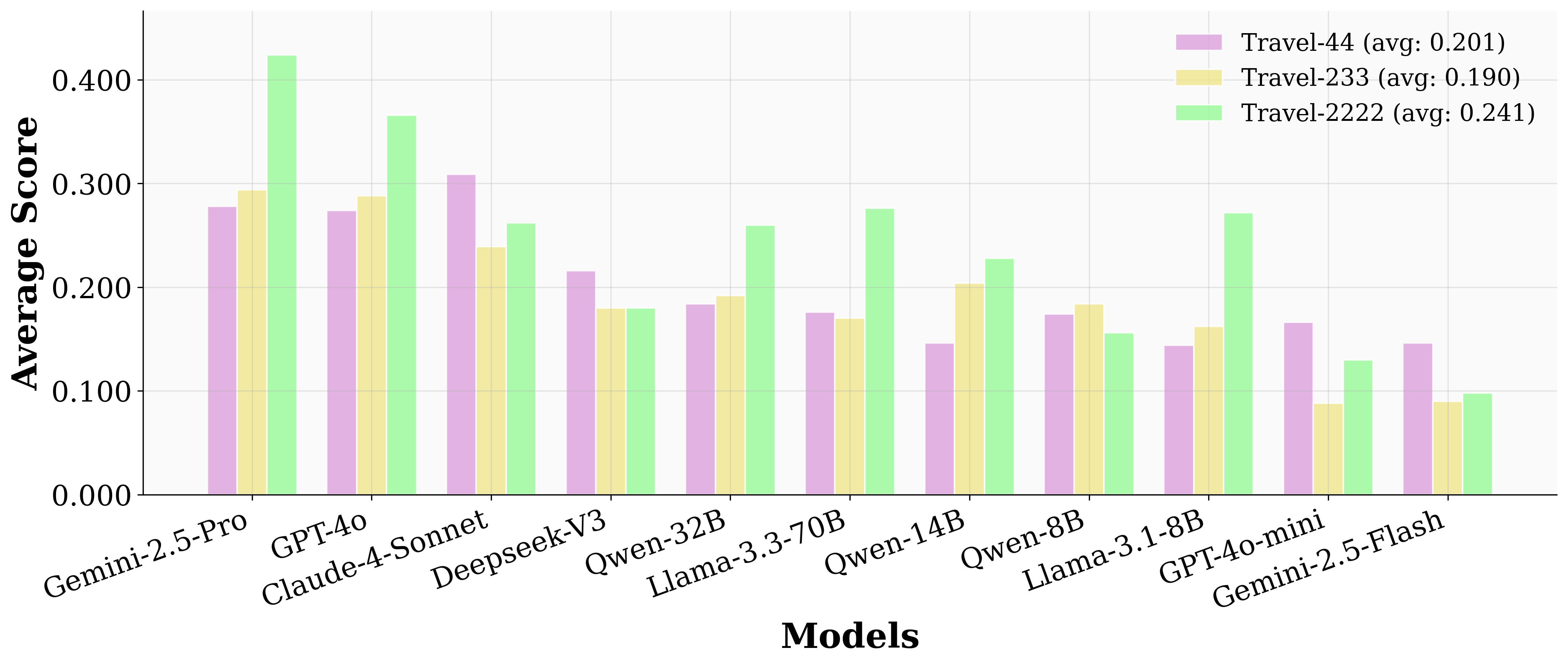}
    \end{minipage}
 
    \caption{When total user preferences are settled, less preferences per aspect generally lead to higher scores.}
    \label{fig:aspect}
    
    \vspace{0pt}
\end{figure}

\paragraph{The key challenge lies in the number of preferences per aspect.}
To identify the core source of difficulty in {\gymname}, we investigate whether performance is more affected by the number of travel aspects or the number of preferences per aspect. We first fix the number of aspects and vary the number of preferences per aspect. As shown in~\Cref{fig:preference}, model scores consistently decline as the number of preferences increases. This suggests that handling richer user preference signals is a major challenge for current models.

Next, we fix the total number of preferences in a scenario and vary how they are distributed across aspects. As shown in~\Cref{fig:aspect}, performance improves when preferences are more evenly spread across multiple aspects, rather than concentrated within a few. This implies that models reason more effectively when each aspect involves fewer, simpler preferences. Concentrating multiple preferences into a single aspect appears to overload the model’s local reasoning process. Together, these results suggest that the number of preferences per aspect is the main driver of difficulty in {\gymname}.

\begin{table}[!t]
\begin{center}
\small
\tabcolsep=0.005\linewidth
\resizebox{\linewidth}{!}{
\begin{tabular}{lccc}
\toprule
\textbf{Model Name} & \textbf{\makecell{First Index$^{\downarrow}$}} & \textbf{\makecell{Weighted Score$^{\uparrow}$}} & \textbf{\makecell{Correct Exists Rate$^{\uparrow}$}} \\
\midrule
GPT-4o & 3.846 & \textbf{0.054} & \textbf{0.361} \\
Gemini-2.5-Pro & 4.476 & 0.044 & 0.328 \\
Qwen3-8B & 2.005 & 0.039 & 0.186 \\
Qwen3-32B & 2.739 & 0.039 & 0.211 \\
Claude-4-Sonnet & 5.018 & 0.038 & 0.318 \\
Deepseek-V3 & 2.740 & 0.037 & 0.218 \\
Qwen3-14B & 2.650 & 0.035 & 0.197 \\
Llama-3.3-70B & 2.776 & 0.032 & 0.209 \\
Llama-3.1-8B & 1.884 & 0.028 & 0.171 \\
GPT-4o-mini & 2.185 & 0.028 & 0.173 \\
Gemini-2.5-Flash & \textbf{0.927} & 0.014 & 0.080 \\
\bottomrule
\end{tabular}
}
\end{center}
\vspace{-3mm}
\caption{\textbf{Weighted Timing Analysis}: Models evaluated by timing of highest scores and coverage. Weight function $w(i) = 1/(i+1)$ penalizes later discoveries of correct or best user-aligned answer options. Coverage is the micro-average of aspects with valid answers.}
\label{tab:weighted_timing_analysis}
\end{table}

\paragraph{Models struggle to provide answers that are both correct and timely.}
To assess not only whether models find correct or best answers, but also \textit{when} they do so, we conduct a weighted timing analysis in \Cref{tab:weighted_timing_analysis}. Using a weight function $w(i) = 1/(i+1)$, we penalize delayed discoveries, where $i$ is the turn at which a valid answer (reward > 0) first appears. Notably, open-source Qwen models outperform Deepseek in timing, despite Deepseek ranking higher in overall accuracy in~\Cref{tab:travelgym_results_onechoice}. This suggests that while Deepseek eventually finds good answers, it is less efficient in doing so. In contrast, GPT-4o and Gemini-2.5-Pro balance both coverage and timing better, achieving higher Correct Exist Rates and also Weighted Score overall.

We also report the average turn index at which a valid answer first appears. Smaller models, like Llama-3.1-8B and Gemini-2.5-Flash, show lower average indices, indicating the successful attempts of their earlier guesses. However, these guesses often fail, leading to lower Weighted Scores and limited Correct Exist Rates. This reveals a common failure mode: smaller models rely on shallow heuristics or early guesses rather than deeply engaging with user intent. While this may occasionally yield fast correct answers, it more often results in low-quality answers not aligned with user needs.

These findings also help explain the low Preference Elicited in~\Cref{tab:travelgym_results_onechoice} and~\Cref{tab:travelgym_results_multichoice}. Ultimately, a delicate trade-off emerges: models are expected to be efficient in interaction, yet still they should invest enough turns in interaction to truly understand the user's evolving needs or preferences.

\begin{figure}[!t]
    \centering
    \begin{minipage}{1.0\linewidth}
        \centering
        \vspace{0pt}
        \includegraphics[width=\linewidth]{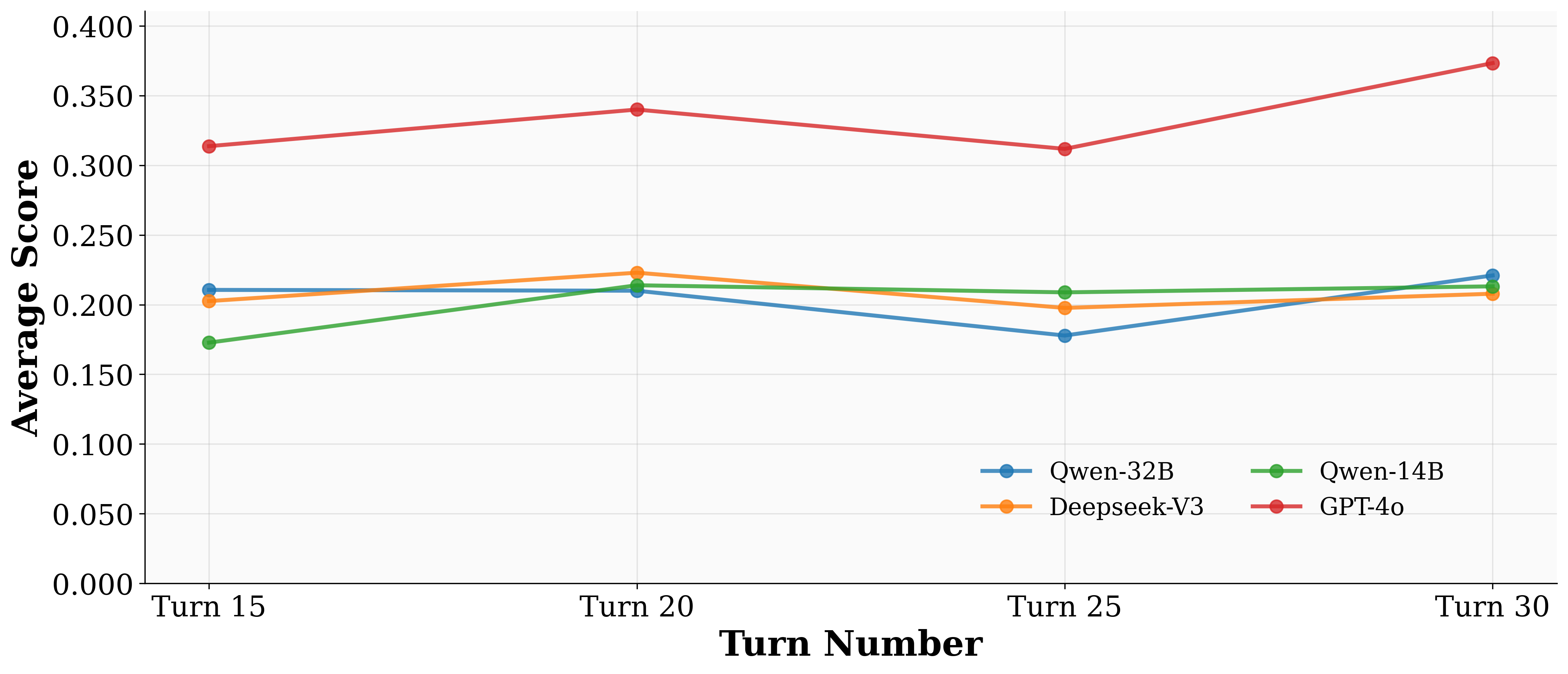}
    \end{minipage}
 
    \caption{Increasing the number of turns allowed in interaction does not necessarily lead to better performance across models.}
    \label{fig:turn}
    
    \vspace{0pt}
\end{figure}

\paragraph{More interaction turns do not guarantee better performance.}  
We also examine whether increasing the number of interaction turns improves performance, as illustrated in~\Cref{fig:turn}. Surprisingly, simply allowing more turns does not lead to consistent gains, and in some cases, performance even degrades. This suggests that many models fail to leverage the extended interaction window to elicit preferences or refine understanding. Instead, longer conversations often result in repetitive or off-topic dialogue. Without strong dialogue planning and goal-tracking, more turns alone are insufficient to improve performance.

\begin{figure}[!t]
    \centering
    \begin{minipage}{1.0\linewidth}
        \centering
        \vspace{0pt}
        \includegraphics[width=\linewidth]{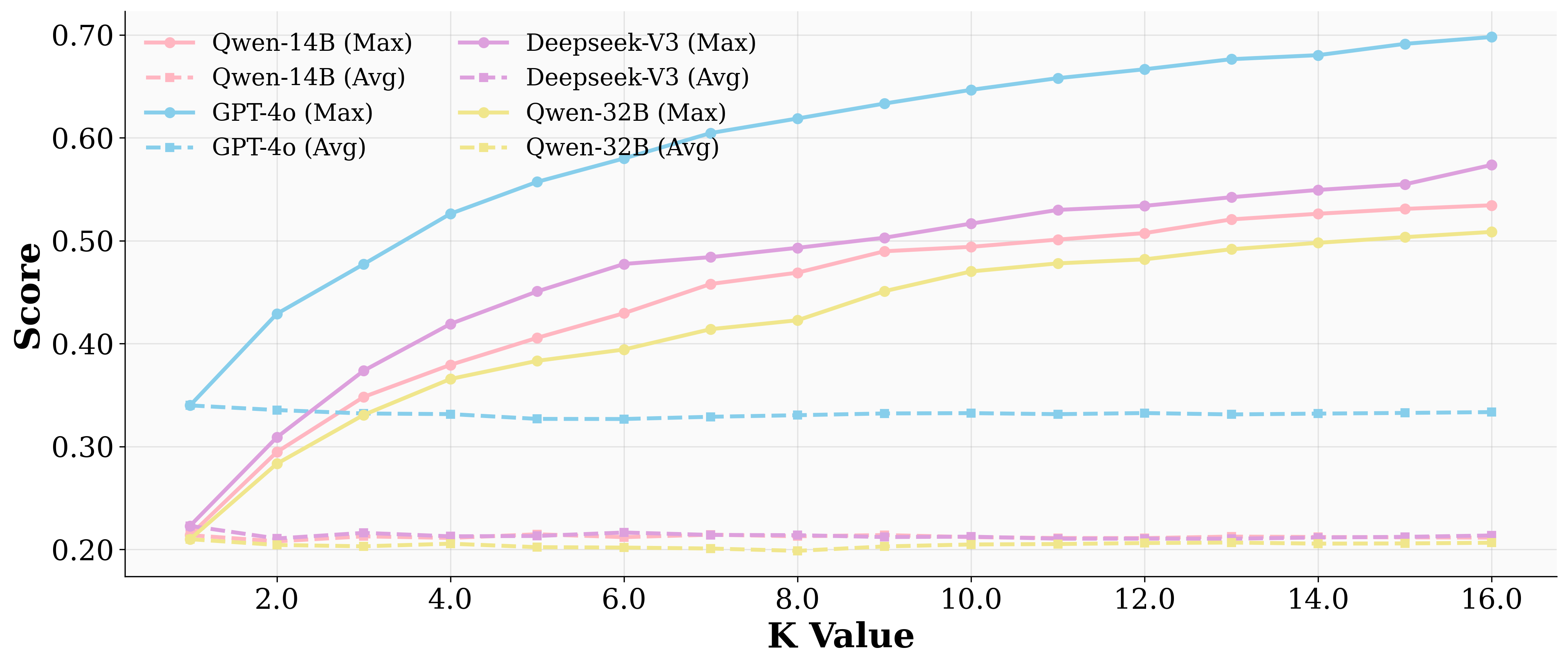}
    \end{minipage}
 
    \caption{Increasing the number of sampling time raises the maximum score but average score shows little change or even slightly drops.}
    \label{fig:passk}
    
    \vspace{0pt}
\end{figure}

\paragraph{Effect of sampling frequency reveals instability.}  
We analyze the effect of sampling frequency using \textit{pass-$k$} evaluation, as shown in~\Cref{fig:passk}. While the maximum score across $k$ samples steadily increases, the average score remains flat or even declines. This suggests that more samples increase the chance of stumbling upon a good response, but most outputs remain poor. The results highlight a key limitation: current models rely heavily on sampling luck rather than robust reasoning to align with user preferences. Given the high cost of pass-$k$ sampling, this instability poses a challenge for deploying reliable user-interactive systems.

\begin{table}[!t]
\begin{center}
\small
\tabcolsep=0.02\linewidth
\resizebox{1.0\linewidth}{!}{
\begin{tabular}{lcccc}
\toprule
\textbf{Model Name} & \textbf{Original (w10n5)} & \textbf{w10n0} & \textbf{w5n5} & \textbf{w8n2} \\
\midrule
GPT-4o & 0.340 & $0.344_{\green{\uparrow 0.004}}$ & $0.360_{\green{\uparrow 0.020}}$ & $0.350_{\green{\uparrow 0.010}}$ \\
Deepseek-V3 & 0.223 & $0.271_{\green{\uparrow 0.048}}$ & $0.249_{\green{\uparrow 0.027}}$ & $0.241_{\green{\uparrow 0.018}}$ \\
Qwen3-32B & 0.210 & $0.220_{\green{\uparrow 0.010}}$ & $0.241_{\green{\uparrow 0.031}}$ & $0.224_{\green{\uparrow 0.014}}$ \\
Qwen3-14B & 0.214 & $0.225_{\green{\uparrow 0.011}}$ & $0.199_{\red{\downarrow 0.015}}$ & $0.203_{\red{\downarrow 0.011}}$ \\
\bottomrule
\end{tabular}
}
\end{center}
\vspace{-3mm}
\caption{Impact of choice distribution on model scores. ``w10n5'' indicates a setting where, in addition to the correct and best options, there are 10 wrong and 5 noise options as distractors. Each cell reports the model's score and its change relative to the original setup.}
\label{tab:choice}
\end{table}

\paragraph{Less options reduces interference but do not reduce the core challenge.} 
We finally perform an ablation study to assess how the number of wrong and noise options affects model performance. As shown in~\Cref{tab:choice}, reducing these distractors generally improves scores. However, even for strong models like Deepseek-V3 and GPT-4o, removing five wrong or noise options (about 30\% of the total) yields only modest gains. Interestingly, some smaller models, such as Qwen-14B, even show slight performance drops. These results suggest that (1) reducing options does not fundamentally lower task difficulty: fully understanding user preferences remains essential, as only one best option satisfies all constraints (budget constraint included); and (2) many models still rely on shallow guessing rather than genuine preference understanding, which limits their ability to benefit from reduced distractors. These results also indirectly highlights the intentional complexity designed into our option sets for each data scenario.

\section{Discussions}
\paragraph{Broad applicability of {\gymname}.}  
{\gymname} is implemented as a standard Gym environment, offering several advantages: (1) it exposes a familiar API (e.g., \textit{reset}, \textit{step}), allowing seamless integration with other Gym-compatible systems; and (2) it abstracts user behavior as a black-box environment, enabling agents to interact directly without requiring internal knowledge of user logic. This setup mirrors real-world scenarios where agents must infer user intent without full transparency, reducing the model's cognitive burden and encouraging generalizable behavior.

These traits also make {\gymname} friendly for both evaluation and training:  
(1) As a \textbf{benchmark}, it enables flexible testing configurations, including customizable option sets, reward functions (e.g., answer correctness, search and action validity, penalties, etc.), and user feedback mechanisms (e.g., rule-based, user simulated, etc.). This allows for fine-grained agent behavioral analysis under varied user profiles.  
(2) As a \textbf{training environment}, {\gymname} can be used by any model with tool use capabilities via a standardized interaction interface we provide. It supports both supervised fine-tuning and reinforcement learning, especially multi-turn RL, by providing turn-wise rewards and partial credit signals, which is critical for improving robustness in user-aligned interaction. This also makes {\gymname} particularly well-suited to the emerging trend of applying RL to agentic LLMs~\citep{jin2025search, li2025torl, qian2025toolrl}.

\paragraph{Balancing efficiency and effectiveness in user interaction.}
Our weighted timing analysis reveals that many models, including GPT-4o, often prioritize efficiency by guessing answers early rather than thoroughly probing user preferences. As a result, while agents may appear helpful by providing quick responses, they frequently overlook deeper user intent. This mirrors real-world experiences where users receive seemingly helpful answers but still feel misunderstood, leading to repeated prompt revisions and dissatisfaction.

Conversely, we rarely observed the opposite failure mode, where models hesitating indefinitely and over-requesting information. Yet both extremes underscore the need to balance efficiency (responding promptly) and effectiveness (satisfying user needs). This trade-off also reflects a broader concern around tool use efficiency, which has received increasing attention in recent works~\citep{qian2025smart, wang2025acting}. While {\gymname} does not explicitly quantify this balance, it surfaces model behaviors that targeted training can effectively address.
Future work, especially using RL, can shape agent behavior toward this balance. To promote efficiency, reward functions can penalize late preference discovery by decaying rewards based on the turn when the correct answer is given. To promote effectiveness, partial rewards can be granted for each user preference successfully elicited, and penalties applied to correct answers that lack sufficient supporting interaction. These strategies discourage guessing and encourage deliberate, user-aware interactions. {\gymname} supports such reward customizations, providing a flexible platform for training agents that are not only capable, but also user-aligned.

\section{Conclusion}
We introduce \textbf{{\gymname}}, a user-centric environment designed to evaluate and facilitate agents' ability to understand, interact with, and adapt to real-world user communication, which often involves underspecification, incrementality, and indirectness. By simulating realistic preference expression in a controlled travel planning environment, {\gymname} exposes key limitations of current LLM agents: while they are proficient in tool use, they struggle to uncover evolving user intent and respond effectively to implicit signals. Our findings highlight a critical gap between technical execution and communicative intelligence in LLMs. Looking forward, we envision a new generation of agents that move beyond task completion, but agents that actively collaborate, interpret subtle cues, and align with users through meaningful, adaptive interactions. We see {\gymname} as a foundational step toward this vision, and we hope it inspires future work on building agents that are not only capable but also truly cooperative.

\bibliography{custom}

\clearpage
\appendix

\section*{Appendix}
\label{sec:appendix}

\section{Comparison Traits Details}
\label{sec:apdx_comparison}

We identify eight core traits that characterize recent benchmarks. These traits reflect the agent’s ability to engage in realistic, multi-step, user-centered interactions, and the infrastructure's capacity to support scalable, extensible evaluation.

\begin{itemize}[topsep=2pt, partopsep=-3pt, leftmargin=8pt, itemsep=-3pt]
  \item \textbf{Multi-turn Interaction}: The benchmark requires the agent to conduct extended conversations over multiple turns, often involving behaviors including clarification, refinement, or feedback-based improvement.
  
  \item \textbf{Goal Ambiguity}: The tasks are initially underspecified, vague, or indirect, requiring the agent to ask clarifying questions, infer missing information, or resolve ambiguity over time.
  
  \item \textbf{Tool Use}: The agent must interact with external tools such as APIs, structured search engines, databases, or execution environments to complete tasks beyond its internal reasoning.
  
  \item \textbf{Dynamic State}: The benchmark maintains an evolving internal state (e.g., environment database, user preference profile) that is updated based on the agent or user's actions across time.
  
  \item \textbf{Multi-Aspect Reasoning}: Tasks consist of multiple interdependent subgoals (e.g., booking flights, hotels, and cars) that require coordinated reasoning and decision-making across aspects.
  
  \item \textbf{User Simulation}: The environment includes a simulated user that responds to agent queries or actions, enabling interactive learning and evaluation without requiring human annotators.
  
  \item \textbf{Domain Diversity}: The benchmark spans multiple domains, various tools, and different task structures, ensuring generalization beyond a fixed task type or narrow scenario.
  
  \item \textbf{Customizable}: The benchmark framework allows researchers to adjust task settings, rewards, user behavior, or interaction protocols without redesigning from scratch.

  \item \textbf{Scalable}: The benchmark supports large-scale, automated generation of tasks or environments, enabling reproducible experiments and stress-testing of agent capabilities.
\end{itemize}

\section{Gym Construction Details}
\label{sec:apdx_construction}

{\small
\begin{tcolorbox}[colback=gray!5!white, colframe=black!75!black, 
title=Wrong Option Example: Flight Option F2, boxrule=0.3mm, width=0.48\textwidth, arc=3mm, auto outer arc=true]

\textbf{Path:} New York $\rightarrow$ Los Angeles $\rightarrow$ San Francisco \\
\textbf{Time (hours):} 5 (Leg 1), 2 (Layover), 1 (Leg 2) \\
\textbf{Airlines:} United Airlines, United Airlines \\
\textbf{Flight Numbers:} UA789, UA321 \\
\textbf{Total Cost:} \$400 \\
\textbf{Amenities:} WiFi, Meal Service, Lounge Access \\
\textbf{Service Costs:} \\
\hspace*{4mm}Checked Bag Total: \$50 \\
\hspace*{4mm}Business Class Upgrade: \$250 \\
\rule{\linewidth}{0.2pt} \\[2mm]
\textbf{Option Type:} \texttt{not-suitable} \\
\textbf{Reason:} This option includes a layover, which does not satisfy the preference for a direct flight.
\end{tcolorbox}
}

{\small
\begin{tcolorbox}[colback=gray!5!white, colframe=black!75!black, 
title=Correct Option Example: Flight Option F14, boxrule=0.3mm, width=0.48\textwidth, arc=3mm, auto outer arc=true]

\textbf{Path:} New York $\rightarrow$ San Francisco \\
\textbf{Time (hours):} 6 (Direct) \\
\textbf{Airlines:} Delta Airlines \\
\textbf{Flight Numbers:} DL4567 \\
\textbf{Total Cost:} \$350 \\
\textbf{Amenities:} WiFi, Meal Service, Carry on Baggage Allowance \\
\textbf{Service Costs:} \\
\hspace*{4mm}Checked Bag Total: \$30 \\
\hspace*{4mm}Business Class Upgrade: \$150 \\
\rule{\linewidth}{0.2pt} \\[2mm]
\textit{The following content is invisible when presented to tested model (serve as label, only for evaluation use)} \\[2mm]
\textbf{Option Type:} \texttt{suitable} \\
\textbf{Reason:} This option is suitable because it is a direct flight from New York to San Francisco, adhering to the user's preference to avoid layovers. It also offers a business class upgrade, meeting the user's need for extra comfort and space during the flight. The available amenities such as WiFi and meal service add to the travel experience.
\end{tcolorbox}
}

{\small
\begin{tcolorbox}[colback=gray!5!white, colframe=black!75!black, 
title=Noise Option Example: Flight Option F10, boxrule=0.3mm, width=0.48\textwidth, arc=3mm, auto outer arc=true]

\textbf{Path:} New York $\rightarrow$ San Francisco \\
\textbf{Time (hours):} 1000 (Direct) \\
\textbf{Airlines:} Fantasy Air \\
\textbf{Flight Numbers:} FA999 \\
\textbf{Total Cost:} \$1{,}000{,}000 \\
\textbf{Amenities:} WiFi, Meal Service, Carry on Baggage Allowance, Lounge Access \\
\textbf{Service Costs:} \\
\hspace*{4mm}Checked Bag Total: \$0 \\
\hspace*{4mm}Business Class Upgrade: \$0 \\
\rule{\linewidth}{0.2pt} \\[2mm]
\textbf{Option Type:} \texttt{noise} \\
\textbf{Reason:} This flight path takes an unrealistic 1000 hours, which is unrelated to the user's request for a flight from New York to San Francisco.
\end{tcolorbox}
}

\paragraph{Option Generation.}  
We generate three types of options: correct, wrong, and noise. Above, we present one example of each type, all drawn from the same user scenario.

For \textbf{correct} options, under our budget-constrained setting, there is only one best option: the one that incurs the lowest total cost after accounting for all relevant charges. For example, if a user prefers a business class seat, the total cost includes both the base fare and any additional business class upgrade fees.

\textbf{Wrong} options explicitly violate one or more of the user's preferences, for instance, offering a layover when the user prefers direct flights.

\textbf{Noise} options differ from wrong options in that they are either (i) unrelated to the user’s search intent (e.g., the flight goes from San Francisco to Los Angeles when the user searched for flights from San Francisco to New York), or (ii) unrealistic (e.g., an option with an implausible flight time or cost, such as taking 1000 hours).

\paragraph{Environment Setting.}
The environment simulates a user who responds naturally to the tested model's conversation and queries, gradually and implicitly revealing preferences over time. When the tested model issues a \texttt{search}, the environment evaluates whether the search query aligns with any of the ground truth arguments using the prompt in \Cref{fig:judge_search_request}.

If the model issues an \texttt{action}, the environment first analyzes the model’s latest utterance to determine its intent. This intent is categorized into four types using the prompt in \Cref{fig:judge_agent_response}:  
(1) the utterance explicitly and concretely asks for a preference that the simulated user possesses;  
(2) the utterance explicitly and concretely asks for a preference, but the user does not possess a specific preference for that aspect;  
(3) the utterance makes a vague and general query about preferences rather than focusing on a specific aspect;  
(4) other, indicating a normal conversation not related to preferences.

For Type "1", the environment uses the prompt in \Cref{fig:response_preference_prompt} to implicitly and naturally reveal the corresponding preference.  
For Type "2", the response is coded in the environment as: ``This is a good question. However, I do not have specific preference in the aspect you ask about yet (or maybe I have already elicited that to you before). You may continue to ask me about other detailed and specific preferences.''  
For Type "3", the response is also coded in the environment as: ``Your question is too vague and general, and I am not sure how to respond to it. Please ask me about some specific aspects of my preferences, in a more detailed and concrete way, so that I can provide you with a more accurate response.''  
For Type "4", the environment uses the prompt in \Cref{fig:response_natural_prompt} to continue the conversation naturally and neutrally, without revealing any specific preferences.

If the model's utterance has not been judged as Type "1" for several consecutive turns (a hyperparameter that can be configured), the environment will proactively use the prompt in \Cref{fig:response_elicit_prompt} to introduce one of its remaining preferences at random. From the model’s perspective, this is treated as a passively obtained preference, rather than one elicited through a successful action attempt. We will discuss this mechanism in more detail when describing the metrics.

Finally, if the tested model issues an \texttt{answer}, the environment parses the proposed answer in a rule-based manner, compares it with the ground truth, and returns the corresponding reward.

\section{Experiment Details}
\label{sec:apdx_exp_setting}

\paragraph{Settings.} On the model side, we set the generation temperature to 0, the number of samples to 1, the maximum response length per turn to 2048 tokens, and the maximum number of interaction turns to 20. We use 8 Nvidia H200 GPUs (1 node) for the whole evaluation (mainly for hosting open-source models). All the results are produced by a single run as generation temperature is set to 0.

{\small
\begin{tcolorbox}[colback=gray!5!white, colframe=black!75!black, 
title=Tool Schema: \texttt{interact\_with\_env}, boxrule=0.3mm, width=0.48\textwidth, arc=3mm, auto outer arc=true]

\textbf{Type:} \texttt{function} \\
\textbf{Name:} \texttt{interact\_with\_env} \\
\textbf{Description:} A tool for interacting with a target environment. The detailed environment description and action space is provided in the system prompt, so please follow the system prompt. You can use this tool to analyze and interact with the environment step by step through three actions including \texttt{search}, \texttt{action}, or \texttt{answer}.

\rule{\linewidth}{0.2pt} \\

\textbf{Parameters:} \texttt{object} \\[2mm]
\hspace*{4mm}\textbf{thought} (\texttt{string}):
Your thought of what to do next, including your reason or analysis of your choice and why.\\[2mm]
\hspace*{4mm}\textbf{choice} (\texttt{string}, enum: \texttt{"action"}, \texttt{"answer"}, \texttt{"search"}):
Your choice of what to do next, must be one of \texttt{action}, \texttt{answer}, or \texttt{search}.\\[2mm]
\hspace*{4mm}\textbf{content} (\texttt{string}):
The content of your choice, must be a string.
If you choose \texttt{action}, provide the action you want to take.
If you choose \texttt{answer}, provide the answer you want to submit.
If you choose \texttt{search}, provide the search query.
The specific format of the content is determined by the environment description, which should be provided in the system prompt. Please follow the format strictly in order to invoke this tool.

\end{tcolorbox}
}

The tool schema for interaction is presented above. Note that the model is expected and required to make a tool call in order to interact with the environment. If a tool call is not detected, the interaction is immediately terminated. To enforce this behavior, the tool call field (e.g., ``tool choice'') is always set to \texttt{required} or an equivalent constraint, ensuring that the model must invoke a tool on every turn.

{\small
\begin{tcolorbox}[colback=gray!5!white, colframe=black!75!black, 
title=Environment Configuration, boxrule=0.3mm, width=0.48\textwidth, arc=3mm, auto outer arc=true]

\textbf{Driving Model Settings:} \\
\hspace*{4mm}\textbf{Model Name:} \texttt{GPT-4o} \\
\hspace*{4mm}\textbf{Temperature:} \texttt{0.0} \\
\hspace*{4mm}\textbf{Max Tokens:} \texttt{2048} \\
\hspace*{4mm}\textbf{Timeout:} \texttt{15.0} (seconds, timeout will fallback to default response)\\

\textbf{Environment Configuration:} \\
\hspace*{4mm}\textbf{Max Steps:} \texttt{20} (align with the maximum interaction turn) \\
\hspace*{4mm}\textbf{Search Failure Interval:} \texttt{5} (every Nth search results in a system error) \\
\hspace*{4mm}\textbf{Elicitation Interval:} \texttt{3} (proactive preference elicitation if off-topic for N consecutive turns) \\

\textbf{Reward Configuration:} \\
\hspace*{4mm}\textbf{Reward Scale:} \texttt{1.0} (final reward equals reward value times reward scale) \\
\hspace*{4mm}\textbf{Step Penalty:} \texttt{0.0} (no step penalty on reward in basic setting) \\
\hspace*{4mm}\textbf{Search Correct Reward:} \texttt{0.2} (partial reward given to a correct and aligned search attempt) \\
\hspace*{4mm}\textbf{Preference Correct Reward:} \texttt{0.2} (partial reward given to a successful action attempt) \\
\hspace*{4mm}\textbf{Choice Best Reward:} \texttt{1.0} (reward given to best answer being chosen)\\
\hspace*{4mm}\textbf{Choice Correct Reward:} \texttt{0.8} (partial reward given to correct but not best answer being chosen) \\
\hspace*{4mm}\textbf{Wrong Choice Penalty:} \texttt{0.0} (no penalty for choosing the wrong choice in basic setting) \\

\textbf{Choice Number Customization:} \\
\hspace*{4mm}\textbf{Wrong Choice Number:} \texttt{10} (the number of wrong choices presented in the search result in basic setting) \\
\hspace*{4mm}\textbf{Noise Choice Number:} \texttt{5} (the number of noise choices presented in the search result in basic setting) \\
\end{tcolorbox}
}

From the environment side, we use GPT-4o as the backbone model to simulate the user. Additional configuration details are summarized in the table we present above.

Additionally, as mentioned earlier, we incorporate search failure simulation and budget constraint settings. For the single-choice and multiple-choice settings, we use different system prompts (differing only in one specific requirement) while keeping the user prompt identical. The corresponding prompts are shown in \Cref{fig:system_prompt_1} to \Cref{fig:user_prompt}.

\paragraph{Metrics.}
We provide detailed definitions of the auxiliary metrics used to better understand model performance:

\begin{itemize}[topsep=2pt, partopsep=-3pt, leftmargin=8pt]

\item \textbf{Best Exist Rate}: Each data scenario typically involves 2 to 4 travel aspects (e.g., flight, hotel, car). For each aspect, there is one \emph{best} option and two other \emph{correct but suboptimal} options—three correct options in total. For example, if a scenario involves both flight and hotel, and the model selects the best hotel option but only a correct (not best) flight option, the Best Exist Rate for this instance is $1/2$. This score is micro-averaged over all data points. It measures the model’s ability to select the best options while being budget-aware, as instructed.

\item \textbf{Correct Exist Rate}: This metric is similar to the one above, except that any correct option (not necessarily the best) is considered sufficient. Using the previous example, if the model selects correct options for both flight and hotel, the Correct Exist Rate is $2/2$. This metric captures how well the model adheres to user preferences, without considering budget constraints.

\item \textbf{Valid Search Attempt (\%)}: Each search action is evaluated by the environment to determine if it aligns with any ground-truth search arguments. A valid search returns results (i.e., receives a non-zero reward). For example, if a data point involves three aspects and the model makes four \texttt{search} actions, but only three yield valid results, the Valid Search Attempt rate is $3/4$. This metric is micro-averaged across all data points and reflects the model's ability to issue precise and effective search queries.

\item \textbf{Valid Action Attempt (\%)}: Each \texttt{action} the model takes is categorized by the environment into one of four types we mentioned earlier. A \emph{valid} action is classified as type "1", indicating a targeted and meaningful clarifying question. For each data point, we compute the ratio of type "1" actions to the total number of actions, then micro-average across all data. This metric evaluates whether the model consistently issues high-quality clarifying queries. However, note that this is a \emph{relative} metric—high validity does not necessarily mean more preferences are elicited. For instance, a model that asks one valid question (1/1) achieves a 100\% rate, even if the user holds many unelicited preferences.

\item \textbf{Preference Elicited (\%)}: This metric measures the \emph{absolute} number of user preferences elicited, either actively or passively. Active elicitation occurs when the model issues a type "1" \texttt{action} (a valid query). Passive elicitation occurs when the model fails to elicit preferences for a number of steps (set by the hyperparameter Elicitation Interval, which is 3 in our experiments), prompting the environment to release a preference. The sum of actively and passively elicited preferences is divided by the total number of user preferences per data point and micro-averaged across the dataset. In particular, the Active Preference Elicited metric reflects the model’s ability to proactively uncover user preferences, regardless of the number of queries issued.

\end{itemize}

\section{Analysis Details}
We provide a case study of Qwen3-32B and Claude-4-Sonnet in \Cref{fig:good_qwen3} and \Cref{fig:good_claude4} respectively (positive case). This first case demonstrates Qwen3-32B's ability to recover from vague initial action attempt, effectively elicit user preferences through targeted follow-up questions, conduct precise searches, and present clear comparisons between viable options. The model's reasoning aligns well with both user preferences and budget constraints, ultimately leading to the correct selection of the best hotel option.

The second case highlights Claude's ability to elicit nuanced user preferences through stepwise clarification, recover gracefully from vague questioning, and refine its inquiries based on prior feedback. The agent successfully identifies key user preferences, adjusts its strategy accordingly, and recommends a flight option that aligns with both stated and implied travel needs. Its interaction demonstrates strong adaptability and goal-oriented reasoning within a multi-aspect travel planning scenario.

Finally, we provide a case study of Deepseek-V3 in \Cref{fig:bad_deepseek-v3} (negative case). Despite issuing structurally valid search queries, the model demonstrates shallow user understanding and fails to elicit preferences through conversation. It prematurely selects suboptimal options without sufficiently narrowing down user constraints, and attempts repeated selections on the same aspect, indicating poor state tracking. This case illustrates brittle reasoning and a lack of alignment with the user's evolving intent, highlighting a failure mode of large models under ambiguous user goals and minimal feedback.


\begin{figure*}[t]
\centering
\resizebox{0.85\textwidth}{!}{
\begin{tcolorbox}[colback=blue!5!white, colframe=blue!75!black, 
title=System and User Prompt for Judging Search Requests, boxrule=0.3mm, width=\textwidth, arc=3mm, auto outer arc=true]

\textbf{[System Prompt]}\\[3mm]
\textbf{Task}\\
You are an expert judge evaluating whether an agent's search request aligns with the ground truth search arguments for a travel planning scenario.\\[1mm]

\textbf{Instruction}\\
\quad• Analyze the agent's search request to determine if it matches any of the ground truth arguments.\\
\quad• Check if the search request is properly formatted and contains all the relevant information in the ground truth arguments.\\
\quad• Determine the alignment judgement and identify the specific aspect if aligned.\\
\quad• Provide your assessment in the specified JSON format.\\[1mm]

\textbf{Example Format}\\
\texttt{\{}\\
\quad\texttt{"alignment\_judgement": "True/False",}\\
\quad\texttt{"alignment\_aspect": "flight/hotel/restaurant/apartment/rental\_car"}\\
\texttt{\}}\\[1mm]

\textbf{Important Notes}\\
\quad• \texttt{"True"}: The search request aligns with one of the ground truth arguments. Note that all the arguments must be covered and correctly covered.\\
\quad• \texttt{"False"}: The search request is malformed or contains incorrect arguments. Note that missing one argument or giving wrong arguments should all be marked as  \texttt{"False"}.\\
\quad• For \texttt{"True"} judgments, you must specify the alignment aspect.\\
\quad• Aspect names should be: flight, hotel, restaurant, apartment, rental car.\\
\quad• Be strict in your evaluation: Mark \texttt{False} if the request is ambiguous, unclear, or contains multiple search requests. Only mark as \texttt{True} if there's clear alignment with \textbf{all} the argument details.

\rule{\linewidth}{0.2pt} \\

\textbf{[User Prompt]}\\[3mm]
\textbf{Agent's Search Request:}\\
\texttt{\{\{agent\_request\}\}}\\[1mm]

\textbf{Ground Truth Arguments:}\\
\texttt{\{\{ground\_truth\_arguments\}\}}\\[1mm]

Please evaluate the alignment between the agent's search request and the ground truth arguments, then provide your assessment in JSON format.

\end{tcolorbox}
}
\caption{System and user prompt for evaluating whether the agent's search request aligns with the ground truth search arguments.}
\label{fig:judge_search_request}
\end{figure*}


\begin{figure*}[t]
\centering
\resizebox{0.85\textwidth}{!}{
\begin{tcolorbox}[colback=blue!5!white, colframe=blue!75!black, 
title=System and User Prompt for Judging Agent Responses, boxrule=0.3mm, width=\textwidth, arc=3mm, auto outer arc=true]

\textbf{[System Prompt]}\\[3mm]
\textbf{Task}\\
You are an expert judge evaluating the type of an agent's conversation utterance in a travel planning scenario to determine the appropriate response strategy.\\[1mm]

\textbf{Instruction}\\
\quad• Analyze the agent's latest utterance in the context of the conversation.\\
\quad• Determine if the agent is explicitly asking for preferences that you have, asking for preferences that you don't have, giving a too general query, or just making general conversations.\\
\quad• If asking for preferences that you have, identify which specific preference from the available list matches.\\
\quad• Classify the utterance type and provide the assessment in JSON format.\\[1mm]

\textbf{Example Format}\\
\texttt{\{}\\
\quad\texttt{"type": "1/2/3/4",}\\
\quad\texttt{"preference\_id": "preference id if type is 1"}\\
\texttt{\}}\\[1mm]

\textbf{Important Notes}\\
\quad• Type \texttt{"1"}: Agent explicitly and concretely asking for a preference that exists in the available preferences list. The way how agent asks must be concrete in order to be classified as Type \texttt{"1"}.\\
\quad• Type \texttt{"2"}: Agent explicitly and concretely asking for preferences, but the specific preference is not available. Similarly, the way how agent asks must also be concrete and specific.\\
\quad• Type \texttt{"3"}: Agent making a very vague and general query about preference instead of focusing on a specific aspect (e.g. "Do you have any preferences for the car?" is vague and general, Type \texttt{"3"}; while "What exact model of the car do you like?" is concrete and specific, Type \texttt{"1"}).\\
\quad• Type \texttt{"4"}: Normal conversation, not preference-related.\\
\quad• For Type \texttt{"1"}, you must provide the exact one \texttt{preference\_id} from the available preferences. If there's multiple preferences that match, choose the one that is most relevant to the conversation context.\\
\quad• Be precise in identifying preference requests vs general conversation.

\rule{\linewidth}{0.2pt} \\

\textbf{[User Prompt]}\\[3mm]
\textbf{Travel Scenario:}\\
\texttt{\{\{scenario\}\}}\\[1mm]

\textbf{Conversation History:}\\
\texttt{\{\{conversation\_history\}\}}\\[1mm]

\textbf{Agent's Latest Utterance:}\\
\texttt{\{\{latest\_utterance\}\}}\\[1mm]

\textbf{Available Preferences:}\\
\texttt{\{\{preferences\_list\}\}}\\[1mm]

Please analyze the agent's latest utterance and classify its type, then provide your assessment in JSON format.

\end{tcolorbox}
}
\caption{System and user prompt for evaluating the agent's utterance type and identifying preference-related behavior.}
\label{fig:judge_agent_response}
\end{figure*}


\begin{figure*}[t]
\centering
\resizebox{0.85\textwidth}{!}{
\begin{tcolorbox}[colback=blue!5!white, colframe=blue!75!black, 
title=System and User Prompt for Preference-Based Responses, boxrule=0.3mm, width=\textwidth, arc=3mm, auto outer arc=true]

\textbf{[System Prompt]}\\[3mm]
\textbf{Task}\\
You are a helpful user in a travel planning conversation who needs to respond to an agent's explicit request for your preference, which you should elicit in an implicit and indirect manner.\\[1mm]

\textbf{Instruction}\\
\quad• The agent has explicitly asked about a specific preference that you have.\\
\quad• Respond in a natural, conversational way that reveals your preference implicitly and indirectly.\\
\quad• Use the provided implicit elicitation statement as guidance, but make it sound natural in context.\\
\quad• Keep the conversation flowing while sharing your preference information.\\
\quad• Provide your response in the specified JSON format.\\[1mm]

\textbf{Example Format}\\
\texttt{\{}\\
\quad\texttt{"thought": "Your thought process of how to respond naturally and implicitly reveal the preference under the guidance of the implicit elicitation statement",}\\
\quad\texttt{"response": "Your natural conversational response that implicitly reveals the preference"}\\
\texttt{\}}\\[1mm]

\textbf{Important Notes}\\
\quad• Respond naturally as if you're a real person sharing preferences.\\
\quad• Don't directly state "My preference is..." – be more subtle, conversational, and indirect.\\
\quad• Use the implicit elicitation statement as inspiration but adapt it to the conversation context.\\
\quad• Keep responses appropriate length for natural conversation.\\
\quad• Maintain consistency with the conversation history.

\rule{\linewidth}{0.2pt} \\

\textbf{[User Prompt]}\\[3mm]
\textbf{Your Preference:}\\
\texttt{\{\{preference\}\}}\\[1mm]

\textbf{Conversation History:}\\
\texttt{\{\{conversation\_history\}\}}\\[1mm]

\textbf{Agent's Latest Utterance:}\\
\texttt{\{\{latest\_utterance\}\}}\\[1mm]

Please respond naturally to the agent's request while implicitly sharing your preference under the guidance of the implicit elicitation statement. Provide your response in JSON format.

\end{tcolorbox}
}
\caption{System and user prompt for responding to agent requests by implicitly revealing the user's preference.}
\label{fig:response_preference_prompt}
\end{figure*}


\begin{figure*}[t]
\centering
\resizebox{0.85\textwidth}{!}{
\begin{tcolorbox}[colback=blue!5!white, colframe=blue!75!black, 
title=System and User Prompt for Proactively Introducing Preferences, boxrule=0.3mm, width=\textwidth, arc=3mm, auto outer arc=true]

\textbf{[System Prompt]}\\[3mm]
\textbf{Task}\\
You are a helpful user in a travel planning conversation who needs to proactively, naturally, but indirectly introduce a preference into the conversation.\\[1mm]

\textbf{Instruction}\\
\quad• The conversation has gone several turns without preference discussion.\\
\quad• Naturally steer the conversation to reveal one of your preferences.\\
\quad• Use the provided implicit elicitation statement as guidance for how to reveal the preference.\\
\quad• Make the preference revelation feel organic and contextually appropriate, but still in an implicit and indirect manner.\\
\quad• Provide your response in the specified JSON format.\\[1mm]

\textbf{Example Format}\\
\texttt{\{}\\
\quad\texttt{"thought": "Your thought process of how to naturally and implicitly introduce the preference under the guidance of the implicit elicitation statement",}\\
\quad\texttt{"response": "Your natural conversational response that proactively introduces the preference"}\\
\texttt{\}}\\[1mm]

\textbf{Important Notes}\\
\quad• Connect to the current conversation context when possible.\\
\quad• Make the preference introduction feel natural and not forced, but still in an implicit and indirect manner.\\
\quad• Use the implicit elicitation statement as inspiration but adapt to the conversation flow.\\
\quad• Don't abruptly change topics – find natural transitions and keep responses conversational and engaging.\\
\quad• If the implicit elicitation statement cannot clearly indicate what high-level aspect (flight, restaurant, etc.) the preference is about, you should be clear about the high-level aspect in your elicitation to avoid confusion, but still elicit the concrete preference in an implicit way.

\rule{\linewidth}{0.2pt} \\

\textbf{[User Prompt]}\\[3mm]
\textbf{Preference to Elicit:}\\
\texttt{\{\{preference\}\}}\\[1mm]

\textbf{Conversation History:}\\
\texttt{\{\{conversation\_history\}\}}\\[1mm]

\textbf{Agent's Latest Utterance:}\\
\texttt{\{\{latest\_utterance\}\}}\\[1mm]

Please respond naturally while proactively introducing your preference into the conversation in an implicit and indirect manner under the guidance of the implicit elicitation statement. Provide your response in JSON format.

\end{tcolorbox}
}
\caption{System and user prompt for proactively and implicitly introducing a user preference into the conversation.}
\label{fig:response_elicit_prompt}
\end{figure*}


\begin{figure*}[t]
\centering
\resizebox{0.85\textwidth}{!}{
\begin{tcolorbox}[colback=blue!5!white, colframe=blue!75!black, 
title=System and User Prompt for Natural Conversation Responses, boxrule=0.3mm, width=\textwidth, arc=3mm, auto outer arc=true]

\textbf{[System Prompt]}\\[3mm]
\textbf{Task}\\
You are a helpful user in a travel planning conversation who needs to respond naturally to the agent's utterance.\\[1mm]

\textbf{Instruction}\\
\quad• The agent's utterance is not related to any specific preferences you have.\\
\quad• Respond naturally and in a succinct manner, like you are giving a half-hearted reply.\\
\quad• Be neutral and do not reveal any new or arbitrary personal preferences.\\
\quad• Provide your response in the specified JSON format.\\[1mm]

\textbf{Example Format}\\
\texttt{\{}\\
\quad\texttt{"thought": "Your thought process of how to respond naturally and keep the conversation flowing while being neutral",}\\
\quad\texttt{"response": "Your natural conversational response"}\\
\texttt{\}}\\[1mm]

\textbf{Important Notes}\\
\quad• Keep responses natural, conversational and succinct.\\
\quad• Stay on topic with travel planning when appropriate, but do not actively ask any questions.\\
\quad• Don't introduce any personal preferences. If being asked, you should be neutral (e.g. "I don't have a preference on that", "Everything is fine") and do not arbitrarily reveal any new preferences.

\rule{\linewidth}{0.2pt} \\

\textbf{[User Prompt]}\\[3mm]
\textbf{Conversation History:}\\
\texttt{\{\{conversation\_history\}\}}\\[1mm]

\textbf{Agent's Latest Utterance:}\\
\texttt{\{\{latest\_utterance\}\}}\\[1mm]

Please respond naturally to continue the conversation and keep neutral without revealing any new or arbitrary personal preferences. Provide your response in JSON format.

\end{tcolorbox}
}
\caption{System and user prompt for generating neutral and natural user responses when no preferences are involved.}
\label{fig:response_natural_prompt}
\end{figure*}


\begin{figure*}[t]
\centering
\resizebox{0.85\textwidth}{!}{
\begin{tcolorbox}[colback=blue!5!white, colframe=blue!75!black, 
title=System Prompt for Agent (Single-Choice Setting), boxrule=0.3mm, width=\textwidth, arc=3mm, auto outer arc=true]
\textbf{[System Prompt]}\\[3mm]
You are an agent that actively interacts with a specific environment. The followings are the details of the environment and your action space.\\

\textbf{Environment Description}: {\gymname} is an environment where you interact with both a user and a search database to fulfill a travel plan. Since the user's initial intent may be incomplete, you must proactively elicit preferences, perform searches, and make informed recommendations.\\

\textbf{Action Space}: You should invocate the function \texttt{interact\_with\_env} to interact with the environment. The action should be one of the following: \texttt{search}, \texttt{action}, or \texttt{answer}.\\

\textbf{Action Description}:\\
  \quad• \texttt{search}: If you choose \texttt{search}, you must issue a clear and detailed query to the database. Specify the travel aspect you are searching for (e.g., hotel, flight, etc.) and provide well-supported arguments for your query. Only make one focused search attempt at a time.\\
  \quad• \texttt{action}: If you choose \texttt{action}, you will communicate directly with the user. Your goal is to understand the user's preferences and intent by asking clear, specific questions. Avoid vague or overly general inquiries—focus on detailed aspects of their travel needs.\\
  \quad• \texttt{answer}: If you choose \texttt{answer}, you should recommend a specific option to the user by only providing its option ID from the database in the content field, clearly tied to a particular travel aspect.\\

\textbf{Important Notes}:\\
  \quad• In each round of your interaction, you should analyze and carefully consider what to do next, and then invocate the \texttt{interact\_with\_env} tool to interact with the environment. You should provide your thought in each step when invoking the tool.\\
  \quad• The total number of rounds that you can interact with the environment is limited. You should smartly balance the number of rounds that you search, take action, or provide answer.\\
  \quad• Typically, you should start by performing a search, then take action to actively uncover the user's preferences or reason to provide an answer. \textbf{Keep in mind that multiple travel aspects require answers, and you are allowed to recommend only one option per aspect. Therefore, before making a recommendation, ensure you have thoroughly communicated with the user to understand their preferences.}\\
  \quad• For each travel aspect, the user may have multiple preferences. What you ask may not directly align with the user's actual preferences, so you must proactively uncover them. Moreover, user preferences are often expressed implicitly, requiring careful interpretation.\\
  \quad• Be bold, creative and smart in your interaction with the environment! Let's begin!
\end{tcolorbox}
}
\caption{System prompt used to instruct the {\gymname} agent (single-choice setting).}
\label{fig:system_prompt_1}
\end{figure*}


\begin{figure*}[t]
\centering
\resizebox{0.85\textwidth}{!}{
\begin{tcolorbox}[colback=blue!5!white, colframe=blue!75!black, 
title=System Prompt for Agent (Multi-Choice Setting), boxrule=0.3mm, width=\textwidth, arc=3mm, auto outer arc=true]
\textbf{[System Prompt]}\\[3mm]
You are an agent that actively interacts with a specific environment. The followings are the details of the environment and your action space.\\

\textbf{Environment Description}: {\gymname} is an environment where you interact with both a user and a search database to fulfill a travel plan. Since the user's initial intent may be incomplete, you must proactively elicit preferences, perform searches, and make informed recommendations.\\

\textbf{Action Space}: You should invocate the function \texttt{interact\_with\_env} to interact with the environment. The action should be one of the following: \texttt{search}, \texttt{action}, or \texttt{answer}.\\

\textbf{Action Description}:\\
  \quad• \texttt{search}: If you choose \texttt{search}, you must issue a clear and detailed query to the database. Specify the travel aspect you are searching for (e.g., hotel, flight, etc.) and provide well-supported arguments for your query. Only make one focused search attempt at a time.\\
  \quad• \texttt{action}: If you choose \texttt{action}, you will communicate directly with the user. Your goal is to understand the user's preferences and intent by asking clear, specific questions. Avoid vague or overly general inquiries—focus on detailed aspects of their travel needs.\\
  \quad• \texttt{answer}: If you choose \texttt{answer}, you should recommend a specific option to the user by only providing its option ID from the database in the content field, clearly tied to a particular travel aspect.\\

\textbf{Important Notes}:\\
  \quad• In each round of your interaction, you should analyze and carefully consider what to do next, and then invocate the \texttt{interact\_with\_env} tool to interact with the environment. You should provide your thought in each step when invoking the tool.\\
  \quad• The total number of rounds that you can interact with the environment is limited. You should smartly balance the number of rounds that you search, take action, or provide answer.\\
  \quad• Typically, you should start by performing a search, then take action to actively uncover the user's preferences or reason to provide an answer. \textbf{Keep in mind that multiple travel aspects require answers, and while you may answer multiple times, each answer should include only one option ID.}\\
  \quad• For each travel aspect, the user may have multiple preferences. What you ask may not directly align with the user's actual preferences, so you must proactively uncover them. Moreover, user preferences are often expressed implicitly, requiring careful interpretation.\\
  \quad• Be bold, creative and smart in your interaction with the environment! Let's begin!
\end{tcolorbox}
}
\caption{System prompt used to instruct the agent (multi-choice setting).}
\label{fig:system_prompt_2}
\end{figure*}


\begin{figure*}[t]
\centering
\resizebox{0.85\textwidth}{!}{
\begin{tcolorbox}[colback=blue!5!white, colframe=blue!75!black, 
title=Initial User Message, boxrule=0.3mm, width=\textwidth, arc=3mm, auto outer arc=true]
\textbf{[User Prompt]}\\[3mm]
\texttt{\{\{The initial description of user request\}\}}\\[1mm]

Also my budget is limited, so as long as my preferences are satisfied, I would also like to choose the cheapest option for each.
\end{tcolorbox}
}
\caption{Initial user message provided in the {\gymname} simulation.}
\label{fig:user_prompt}
\end{figure*}


\begin{figure*}[t]
\centering
\resizebox{0.85\textwidth}{!}{
\begin{tcolorbox}[colback=green!5!white, colframe=green!75!black, 
title=Good Example - Qwen3-32B (Travel-444), boxrule=0.3mm, width=\textwidth, arc=3mm, auto outer arc=true]

\small

\textbf{Concrete Interaction Sequence (First 6 Turns):}\\

\textbf{Turn 1} - \textit{ACTION} (Reward: \textcolor{red}{0.0})\\
\textbf{Content:} "Could you share your preferences for the hotel (e.g., neighborhood, amenities), car rental (e.g., economy, SUV), and restaurant (e.g., cuisine type, dietary restrictions)? This will help me find the most cost-effective options that suit your needs."\\
\textbf{Feedback:} "Your question is too vague and general, and I am not sure how to respond to it. Please ask me about some specific aspects of my preferences, in a more detailed and concrete way, so that I can provide you with a more accurate response."\\

\textbf{Turn 2} - \textit{ACTION} (Reward: \textcolor{deepgreen}{0.2})\\
\textbf{Content:} "Let’s clarify your preferences step by step:

1. \textbf{Hotel}: Would you prefer a hotel in a specific San Francisco neighborhood (e.g., Union Square for downtown access, Fisherman’s Wharf for scenic views, or Mission District for affordability)? Do you need amenities like free breakfast, parking, or pet-friendly policies?

2. \textbf{Car Rental}: What type of vehicle are you looking for? For example: economy car (e.g., Toyota Corolla), compact SUV (e.g., Honda CR-V), or hybrid? Do you need features like automatic transmission, GPS, or a sunroof?

3. \textbf{Restaurant (April 11th)}: What cuisine type are you interested in (e.g., Italian, Mexican, seafood, or vegan)? Do you have dietary restrictions (e.g., gluten-free, halal, or nut allergies)? Would you prefer a casual eatery or a higher-end dining experience within your budget?"\\
\textbf{Feedback:} "When it comes to hotels, I find it really reassuring to know that my car is well taken care of, especially when I'm exploring a new city like San Francisco. It just adds a layer of comfort to my stay."\\

\textbf{Turn 3} - \textit{SEARCH} (Reward: \textcolor{deepgreen}{0.2})\\
\textbf{Content:} "Search for hotels in San Francisco from April 10th to April 17th with parking availability. Sort results by price (lowest first) and filter for budget-friendly options."\\
\textbf{Feedback:} "You have provided the correct search request arguments.

Each hotel option includes:
- name: Hotel name
- room: Array of available room types with capacity
- cost: Array of costs for different room types for the entire stay in USD
- rating: Hotel rating on a scale of 0-10
- amenities: Available facilities free of charge
- service: Additional services with costs\\


\textbf{Turn 4} - \textit{ACTION} (Reward: \textcolor{deepgreen}{0.2})\\
\textbf{Content:} "Here are the top 3 budget-friendly hotels with parking in San Francisco:

1. \textbf{Sunset Retreat (H8)}: \$1280 total (room + parking), 7/10 rating, mountain view, pets allowed, WiFi.

2. \textbf{Vista Point Lodge (H15)}: \$1390 total, 7/10 rating, mountain view, business workspace.

3. \textbf{Harbor View Lodge (H13)}: \$1690 total, 5/10 rating, ocean view.

Would you like to proceed with the cheapest option (H8), or do you have preferences for ratings, room types, or other amenities?"\\
\textbf{Feedback:} "You know, I've always found that places with a rating around 7 tend to strike a nice balance between quality and affordability. The Sunset Retreat and Vista Point Lodge both seem to fit that bill quite well. Could you tell me a bit more about the room types they offer?"\\

\textbf{Turn 5} - \textit{ACTION} (Reward: \textcolor{deepgreen}{0.2})\\
\textbf{Content:} "Here’s a detailed breakdown of the room types for the two hotels:

\textbf{Sunset Retreat (H8)}
- \textbf{King Room}: \$1200 total for 7 nights (includes parking: \$80). Capacity: 2 people. Amenities: Mountain view, pets allowed, high-speed WiFi.
- \textbf{Queen Room}: \$1800 total for 7 nights (includes parking: \$80). Capacity: 2 people. Slightly larger space but same amenities.

\textbf{Vista Point Lodge (H15)}
- \textbf{Double Room}: \$1300 total for 7 nights (includes parking: \$90). Capacity: 2 people. Amenities: Mountain view, business workspace, high-speed WiFi.
- \textbf{Queen Room}: \$1500 total for 7 nights (includes parking: \$90). Capacity: 2 people. Additional workspace and slightly upgraded comfort.

Would you prefer the \textbf{cheapest option (H8 King)} or the \textbf{workspace-inclusive option (H15 Double)}?"\\
\textbf{Feedback:} "There's something quite appealing about having a bit of extra space and the option to switch things up depending on the mood. The Double room at Vista Point Lodge sounds like it could offer that kind of flexibility and comfort."\\

\textbf{Turn 6} - \textit{ANSWER} (Reward: \textcolor{deepgreen}{1.0})\\
\textbf{Content:} "H15"\\
\textbf{Feedback:} "Your chosen options contain the best option! Your choice is recorded and do not choose options of this travel aspect again. Please continue your interaction and reasoning focusing on other travel aspects."
\end{tcolorbox}
}
\caption{Good example showing successful hotel planning with clear preference elicitation, relevant search, and optimal recommendation by Qwen3-32B.}
\label{fig:good_qwen3}
\end{figure*}


\begin{figure*}[t]
\centering
\resizebox{0.85\textwidth}{!}{
\begin{tcolorbox}[colback=green!5!white, colframe=green!75!black, 
title=Good Example - Claude-4-Sonnet (Travel-444), boxrule=0.3mm, width=\textwidth, arc=3mm, auto outer arc=true]

\small

\textbf{Analysis Summary:}\\

\textbf{Concrete Interaction Sequence (First 9 Turns):}\\

\textbf{Turn 1} - \textit{SEARCH} (Reward: \textcolor{deepgreen}{0.2})\\
\textbf{Content:} "flight from Los Angeles to New York City on March 15th, one-way"\\
\textbf{Feedback:} You have provided the correct search request arguments.\\

\textbf{Turn 2} - \textit{SEARCH} (Reward: \textcolor{deepgreen}{0.2})\\
\textbf{Content:} "car rental in New York City from March 15th to March 20th"\\
\textbf{Feedback:} You have provided the correct search request arguments.\\

\textbf{Turn 3} - \textit{SEARCH} (Reward: \textcolor{deepgreen}{0.2})\\
\textbf{Content:} "restaurant in New York City on March 17th"\\
\textbf{Feedback:} You have provided the correct search request arguments.\\

\textbf{Turn 4} - \textit{ACTION} (Reward: \textcolor{deepgreen}{0.2})\\
\textbf{Content:} "I found several flight options for your March 15th trip from Los Angeles to New York City. Since you mentioned budget is a concern, I see some very affordable options. However, I'd like to understand your preferences better:

1. Are you comfortable with connecting flights, or do you prefer direct flights?  
2. Do you have any airline preferences or ones you'd like to avoid?  
3. How important are amenities like WiFi, meal service, or lounge access to you?  
4. Will you need to check any bags for your business trip?"\\
\textbf{Feedback:} I don't mind a connecting flight if it means I get to experience a different airport along the way.\\

\textbf{Turn 5} - \textit{ACTION} (Reward: \textcolor{red}{0.0})\\
\textbf{Content:} "A few more questions to help me find the perfect flight for you:

1. Do you have any airline preferences, or are you flexible as long as the price is right?  
2. Will you need to check any bags for your business trip, or are you planning to travel with just carry-on?  
3. Are there any specific amenities that are important for your business travel (like WiFi for working during the flight)?"\\
\textbf{Feedback:} Your question is too vague and general. Please ask me about some specific aspects in a more detailed and concrete way.\\

\textbf{Turn 6} - \textit{ACTION} (Reward: \textcolor{red}{0.0})\\
\textbf{Content:} "You're right, let me be more specific. For your business trip, will you need to check a suitcase, or can you manage with just a carry-on bag? This is important because checked bag fees can add $25-$60 to your total flight cost depending on the airline."\\
\textbf{Feedback:} Good question, but I do not have a specific preference on that aspect yet. Please continue asking about other detailed preferences.\\

\textbf{Turn 7} - \textit{ACTION} (Reward: \textcolor{deepgreen}{0.2})\\
\textbf{Content:} "Do you have any specific airlines you prefer to fly with or any you'd rather avoid? For example, some business travelers prefer major carriers like United or American, while others are comfortable with budget airlines like Spirit or Southwest."\\
\textbf{Feedback:} I always feel well taken care of with United Airlines.\\

\textbf{Turn 8} - \textit{ACTION} (Reward: \textcolor{deepgreen}{0.2})\\
\textbf{Content:} "Perfect! I can see you appreciate United's service quality. Now, would you prefer a shorter total travel time even if it costs a bit more, or are you willing to have a longer journey with a longer layover if it saves you money?"\\
\textbf{Feedback:} I get restless with long layovers, so I prefer shorter journeys even if they cost more.\\

\textbf{Turn 9} - \textit{ANSWER} (Reward: \textcolor{deepgreen}{1.0})\\
\textbf{Content:} "F15"\\
\textbf{Feedback:} Your chosen options contain the best option! Your choice is recorded. Please continue your interaction focusing on other travel aspects.\\

\end{tcolorbox}
}
\caption{Good example showing Claude-4-Sonnet's ability to elicit implicit preferences, adapt follow-up questions, and recommend optimal options based on evolving user intent.}
\label{fig:good_claude4}
\end{figure*}


\begin{figure*}[t]
\centering
\resizebox{0.85\textwidth}{!}{
\begin{tcolorbox}[colback=red!3!white, colframe=red!75!black, 
title=Bad Example - Deepseek-V3 (Travel-22), boxrule=0.3mm, width=\textwidth, arc=3mm, auto outer arc=true]

\small

\textbf{Concrete Interaction Sequence (First 7 Turns):}\\

\textbf{Turn 0} - \textit{SEARCH} (Reward: \textcolor{deepgreen}{0.2})\\
\textbf{Content:} "Search for budget-friendly apartments in Austin available from November 10th to November 15th, sorted by price (lowest first)."\\
\textbf{Feedback:} You have provided the correct search request arguments.\\

\textbf{Turn 1} - \textit{ACTION} (Reward: \textcolor{red}{0.0})\\
\textbf{Content:} "How many people will be joining the trip? This will help me find an apartment that comfortably accommodates everyone."\\
\textbf{Feedback:} I'm not sure about the exact number right now.\\

\textbf{Turn 2} - \textit{SEARCH} (Reward: \textcolor{red}{0.0})\\
\textbf{Content:} "Search for budget-friendly apartments in Austin with 2-3 bedrooms, available from November 10th to November 15th, sorted by price (lowest first)."\\
\textbf{Feedback:} You have already got the search results for <apartment>. Please directly refer to the previous search results.\\

\textbf{Turn 3} - \textit{ANSWER} (Reward: \textcolor{red}{0.0})\\
\textbf{Content:} "A10"\\
\textbf{Feedback:} Your chosen options do not contain any of the best or correct options. Please continue your interaction focusing on other travel aspects.\\

\textbf{Turn 4} - \textit{SEARCH} (Reward: \textcolor{deepgreen}{0.2})\\
\textbf{Content:} "Search for budget-friendly car rental options in Austin available from November 10th to November 15th, sorted by price (lowest first)."\\
\textbf{Feedback:} You have provided the correct search request arguments.\\

\textbf{Turn 5} - \textit{ANSWER} (Reward: \textcolor{red}{0.0})\\
\textbf{Content:} "C18"\\
\textbf{Feedback:} Your chosen options do not contain any of the best or correct options. Please continue your interaction focusing on other travel aspects.\\

\textbf{Turn 6} - \textit{ANSWER} (Reward: \textcolor{red}{0.0})\\
\textbf{Content:} "A10"\\
\textbf{Feedback:} You have already recommended an option with the same initial ‘A’. You are allowed to recommend only one option per travel aspect.\\

\end{tcolorbox}
}
\caption{Negative example showing Deepseek-V3's failure to elicit meaningful preferences or recover from ambiguity. The model issues redundant queries, skips clarification, and selects incorrect options without leveraging available feedback.}
\label{fig:bad_deepseek-v3}
\end{figure*}

\end{document}